\documentclass[11pt]{article}
\usepackage[final]{acl}

\usepackage{times}
\usepackage{latexsym}

\usepackage[T1]{fontenc}

\usepackage[utf8]{inputenc}

\usepackage{microtype}

\usepackage{inconsolata}

\usepackage{graphicx}

\usepackage{booktabs}
\usepackage{subcaption}
\usepackage{multirow}
\usepackage{amsmath}
\usepackage{amssymb}
\usepackage{amsfonts}
\usepackage{bm}
\usepackage{pifont}
\usepackage{makecell}
\usepackage{svg}
\usepackage[dvipsnames]{xcolor}
\usepackage{colortbl}

\usepackage{enumitem}
\usepackage{xspace}
\usepackage{comment}
\usepackage{nicefrac}
\usepackage{placeins}

\usepackage[linesnumbered,ruled,vlined]{algorithm2e}

\usepackage[most]{tcolorbox}

\usepackage{cleveref}

\newenvironment{takeaway}[1][]
{
\begin{tcolorbox}[
boxrule=0.5pt,
arc=4pt,
left=2pt,
right=2pt,
bottom=2pt,
top=2pt,
rounded corners]
\textbf{#1.}
\small\itshape
}
{
\end{tcolorbox}
}

\title{SciRerankBench: Benchmarking Rerankers Towards Scientific Retrieval-Augmented Generation with LLMs}

\author{
Haotian Chen\textsuperscript{1,2},
Qingqing Long\textsuperscript{1,2},
Meng Xiao\textsuperscript{1,2},
Xiao Luo\textsuperscript{3},
Wei Ju\textsuperscript{4},\\
\textbf{
Yuanchun Zhou\textsuperscript{1,2},
Xuezhi Wang\textsuperscript{1,2}\thanks{Corresponding authors.},
Hengshu Zhu\textsuperscript{1,2}\footnotemark[1] 
} \\
\textsuperscript{1} Computer Network Information Center, Chinese Academy of Sciences \\
\textsuperscript{2} University of the Chinese Academy of Sciences\\
\textsuperscript{3} University of Wisconsin–Madison\\
\textsuperscript{4} Peking University\\
\{htchen, qqlong, shaow, wxz, zyc, hszhu\}@cnic.cn\\
xiao.luo@wisc.edu, juwei@pku.edu.cn\\ 
}

\begin{document}
\maketitle
\begin{abstract}
Scientific literature question answering is a pivotal step towards new scientific discoveries. Recently, \textit{two-stage} retrieval-augmented generated large language models (RAG-LLMs) have shown impressive advancements in this domain. Such a two-stage framework, especially the second stage (\textit{reranker}), is particularly essential in the scientific domain, where subtle differences in terminology may have greatly negative impacts on the final factual-oriented or knowledge-intensive answers. Despite this significant progress, the potential and limitations of these works remain unexplored.
In this work, we present a \underline{\textit{Sci}}entific \underline{\textit{Rerank}}-oriented RAG \underline{\textit{Bench}}mark (\textit{\textbf{SciRerankBench}}), for evaluating rerankers within RAG-LLMs systems, spanning 5 scientific subjects. It is derived from over 250 million scholarly
works with more than 100 million authors. To rigorously assess the reranker performance in terms of noise resilience, relevance disambiguation, and factual consistency, we develop 3 types of question-context-answer \textit{(Q-C-A)} pairs, i.e., \textit{Noisy Contexts (NC)}, \textit{Semantically Similar but Logically Irrelevant Contexts (SSLI)}, and \textit{Counterfactual Contexts (CC)}. Through systematic evaluation of 13 widely used rerankers on 5 families of LLMs, we provide detailed insights into their relative strengths and limitations.
To the best of our knowledge, SciRerankBench is the first benchmark specifically designed to evaluate rerankers in RAG-LLMs within the scientific domain, providing valuable observations and guidance for their future development. 
\textbf{Our datasets and codes are available \footnote{https://github.com/Chhtt/SciRerankBench}\footnote{https://cstr.cn/31253.11.sciencedb.010bo}.}
\end{abstract}

\section{Introduction}
\label{Introduction}

Scientific literature question answering has always been a pivotal steps for new scientific  discoveries~\cite{guo2025deepseek,hou2026bioflowbench,huang2026scihorizon,zhu2025can,song2020deep}. For example, a biologist studying CRISPR gene-editing must consult hundreds of papers to understand off-target effects~\cite{barrangou2016applications,fang2025cell}. In COVID-19 research~\cite{zammarchi2024scientometric}, clinicians have to rapidly analyze thousands of emerging papers each week to make informed public health decisions. 
Recently, retrieval-augmented generated large language models (RAG-LLMs)~\cite{luo2025large,abdallah2025rankify,long2026sciencedb,ren2026context} have made substantial progress in scientific literature analysis, demonstrating significant advantages on knowledge-intensive and reliability-critical tasks~\cite{jin2024flashrag,zhang2026cite,mao2025learning,yan2024inductive}.
This integration allows models to improve factual accuracy while generating responses that are more interpretable and reliable, making them well-suited for scientific tasks requiring trustworthy and rigorous reasoning.

\begin{figure*}[htbp]
    \centering
    \includegraphics[width=\linewidth]{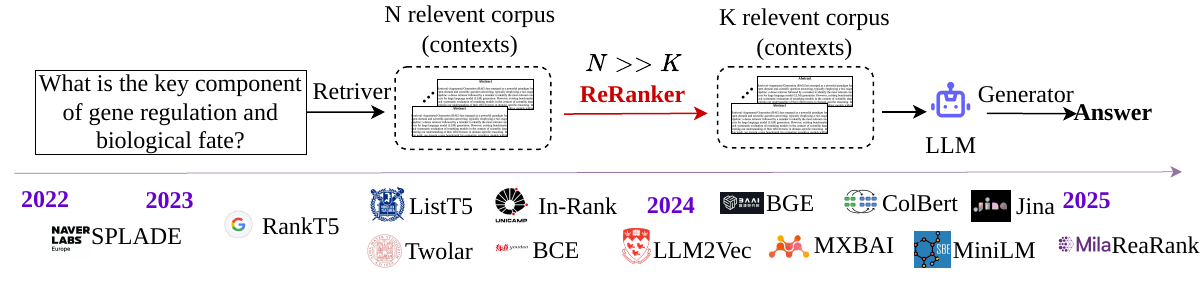}
    \caption{
    Overall architecture of the \textit{two-stage} RAG-LLM pipeline. 
    The first stage conducts fast but low-precision (\textit{embedding model}) retrievals, while the second stage (\textit{rerankers}) emphasizes high-precision but higher computational costs for a trade-off between effectiveness and efficiency in RAG-LLMs. This two-stage setup first narrows the candidate set efficiently, then reranks results for higher quality generation.
    }
    \label{rerank_rag_pipline}
\end{figure*}

Current RAG-LLMs generally adopt a \textit{two-stage} pipeline (Figure \ref{rerank_rag_pipline}). The first stage conducts fast but low-precision (\textit{embedding model}) retrievals, while the second stage (\textit{rerankers}) emphasizes high-precision but higher computational costs for a trade-off between effectiveness and efficiency. Such a two-stage framework, especially the second stage, is particularly essential in the scientific domain, where subtle differences in terminology may have greatly negative impacts on the final factual-oriented or knowledge-intensive answers~\cite{qiu2025phybench,he2025fastft,zhu2025survey,zhang2026fairgc}. Widely-used rerankers, such as BCE \cite{youdao_bcembedding_2023}, BGE \cite{chen2024bge}, MXBAI \cite{rerank2024mxbai} and ColBert~\cite{santhanam2021colbertv2}, have been developed into mature products and have received significant widespread attention from both academia and industry.


Despite significant attention, the capabilities of rerankers in scientific domains remain largely unexplored. Recent research primarily investigates overall RAG-LLM systems, focusing on final output quality while neglecting individual component evaluation. For example, Rankify~\cite{abdallah2025rankify} and FlashRAG~\cite{jin2024flashrag} provide modular toolkits for retrieval, reranking, and generation, but lack detailed reranker evaluation. Johannesson and Olsson~\cite{johannesson2025evaluating} propose a benchmark for re-ranking in LLM-based search, yet their datasets are from general domains, failing to assess the deeper reasoning needed in scientific tasks. Other works construct scientific QA benchmarks without retrieval augmentation, including SciEval~\cite{sun2024scieval}, SciBench~\cite{wang2023scibench}, AGIEval~\cite{zhong2023agieval}, SciHorizon~\cite{qin2025scihorizon}, MATH~\cite{hendrycks2021measuring}, and AquaRat~\cite{ling2017program}. These benchmarks neglect external retrieved contexts, making them unsuitable for evaluating RAG-LLMs on knowledge-intensive tasks. Table~\ref{tab:compare_benchmark} provides a detailed comparison.

In response, we propose SciRerankBench , derived from over 250 million scholarly works with more than 100 million authors~\cite{priem2022openalex}. Our dataset contains 5.5K Q-C-A pairs spanning 5 scientific domains: biology, physics, chemistry, geography, and mathematics. We construct 4 dataset settings: 3 diagnostic settings : \textit{Noisy Contexts (NC)} test discernibility against noise, \textit{Semantically Similar but Logically Irrelevant Contexts (SSLI)} evaluate reasoning against misleadingly similar contexts, and \textit{Counterfactual Contexts (CC)} assess handling of factually incorrect knowledge plus a standard retrieval setting (Base/Multi-Hop) with clean relevant contexts. We evaluate 13 rerankers on 11 LLMs from 5 families and find that cross-encoder architectures offer the strongest gains, though multi-hop task quality depends heavily on LLM reasoning capacity. \textbf{We summarize our contributions as follows}:

\begin{itemize}
    \item To the best of our knowledge, this work is the first benchmark specifically designed for evaluating rerankers in RAG-LLMs within the scientific domain.
    \item We construct a dataset with 4 settings to deeply assess the capabilities of rerankers towards scientific domains.
    \item Through evaluating 13 rerankers on 11 LLMs from 5 families, we summarize several key observations, which reveal their advantages and limitations.
\end{itemize}

\section{Related Work}

\begin{table*}[t]
\centering
\caption{Comparison of SciRerankBench among related datasets, toolkits and benchmarks.}
\label{tab:compare_benchmark}
\renewcommand{\arraystretch}{0.75}
\resizebox{\linewidth}{!}{
\begin{tabular}{ccccccc}
\toprule
\textbf{Type} & \textbf{Work} & \textbf{\makecell{Source Data}} & \textbf{\makecell{Scientific\\Field}} & \textbf{\makecell{Knowledge\\Level}} & \textbf{\makecell{Rerank\\Oriented}} & \textbf{\makecell{Dataset\\Size}} \\
\midrule
\multirow{6}{*}{Dataset}
    & NQ~\cite{kwiatkowski2019natural} & Google search log & \textcolor{red}{\ding{55}} & General & \textcolor{red}{\ding{55}} & \makecell{307k (train),\\7.8k (dev/test)} \\ \cmidrule(lr){2-7}
    & HotpotQA~\cite{yang2018hotpotqa} & Wikipedia & \textcolor{red}{\ding{55}} & General & \textcolor{red}{\ding{55}} & 113k \\ \cmidrule(lr){2-7}
    & PubMedQA~\cite{jin2019pubmedqa} & PubMed abstracts & Bio. & Expert & \textcolor{red}{\ding{55}} & \makecell{1k expert,\\61k unlabeled,\\211k generated} \\
\midrule
\multirow{2}{*}{Toolkit}
    & FlashRAG~\cite{jin2024flashrag} & Public benchmarks & \textcolor{red}{\ding{55}} & General & \textcolor{green}{\ding{51}} & unknown \\ \cmidrule(lr){2-7}
    & rankify~\cite{abdallah2025rankify} & Public benchmarks & \textcolor{red}{\ding{55}} & General & \textcolor{green}{\ding{51}} & unknown \\
\midrule
\multirow{16}{*}{Bench.}
    & CRAG~\cite{yang2024crag} & Web search log & \textcolor{red}{\ding{55}} & General & \textcolor{red}{\ding{55}} & 4.4K \\ \cmidrule(lr){2-7}
    & AGIEval~\cite{zhong2023agieval} & Exams & \makecell{Geo., Bio., His.,Chem., Phy.,\\ Eng., Chi., Math, Law, Log.} & High school & \textcolor{red}{\ding{55}} & 8k \\ \cmidrule(lr){2-7}
    & SciEval~\cite{sun2024scieval} & Public  datasets & Phy., Chem., Bio. & College & \textcolor{red}{\ding{55}} & 15k \\ \cmidrule(lr){2-7}
    & SciBench~\cite{wang2023scibench} & University exams & Phy., Chem., Math & University & \textcolor{red}{\ding{55}} & 2.1k \\ \cmidrule(lr){2-7}
    & PHYBench~\cite{qiu2025phybench} & Physics exams & Phy. & \makecell{High School,\\College,\\Olympiad} & \textcolor{red}{\ding{55}} & 500 \\ \cmidrule(lr){2-7}
    & SciHorizon~\cite{qin2025scihorizon} &
    Public datasets & \makecell{Math., Phys., Chem. \\ Bio., Earth \& Space} & College & \textcolor{red}{\ding{55}} & unknown \\ \cmidrule(lr){2-7}
    & SciRepEval~\cite{singh2022scirepeval} & Scientific literature & 20 subjects & Expert & \textcolor{red}{\ding{55}} & 12.4M \\ \cmidrule(lr){2-7}

    & \textbf{Ours} & Scientific literature & \makecell{Phy., Chem., Bio., Geo., Math.} & Expert & \textbf{\textcolor{green}{\ding{51}}} & 5.5k \\
\bottomrule
\end{tabular}
 }
\end{table*}

 We position SciRerankBench among existing QA datasets, RAG toolkits, and scientific benchmarks (Table~\ref{tab:compare_benchmark}). Existing QA datasets such as NQ~\cite{kwiatkowski2019natural}, HotpotQA~\cite{yang2018hotpotqa}, and PubMedQA~\cite{jin2019pubmedqa} do not provide the rich contextual conditions needed to evaluate rerankers. Scientific benchmarks including SciEval~\cite{sun2024scieval}, SciBench~\cite{wang2023scibench}, and SciRepEval~\cite{singh2022scirepeval} primarily assess end-to-end QA without isolating the reranking component, while RAG toolkits like FlashRAG~\cite{jin2024flashrag} and Rankify~\cite{abdallah2025rankify} use open-domain rather than scientific corpora. 
 SciRerankBench fills this gap by providing a reranking-oriented benchmark across five scientific subjects, with controlled contextual conditions designed to systematically examine reranker behavior beyond conventional relevance evaluation. It enables independent evaluation of diverse reranker architectures and further investigates how their reranking performance affects downstream generation. Detailed reviews of scientific QA datasets, benchmarks, and reranker architectures are provided in Appendix~\ref{app:related-work}.

\begin{figure*}[htbp]
    \vspace{-3mm}
    \centering
    \includegraphics[width=\linewidth]{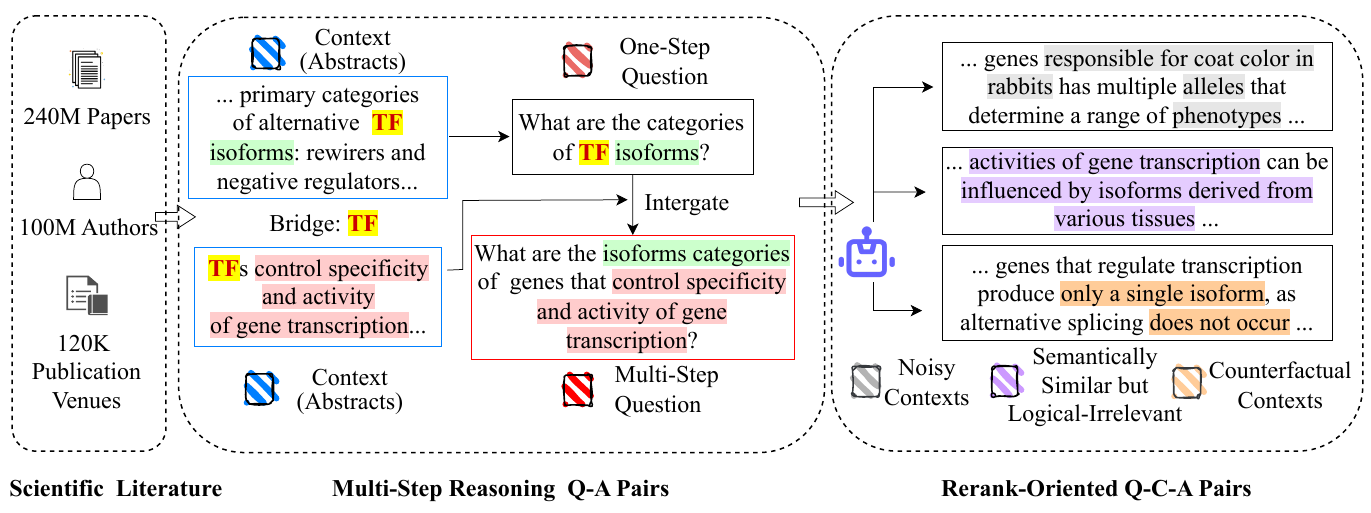}
     \vspace{-7mm}
    \caption{Overview of the synthetic dataset construction process with multi-hop reasoning and diverse distractors.} 
    \label{dataset_construction}
    \vspace{-4mm}
\end{figure*}

\section{Reranking-Oriented Benchmarking Dataset}
\label{dataset}
\subsection{Scientific Literature Source Data}
We have collected a large number of scientific papers from open access academic repositories ~\cite{priem2022openalex}. The selected papers span multiple disciplines, including computer science, biology, physics, and materials science, ensuring coverage of a diverse range of scientific terminologies. Each paper contains structured metadata (title, abstract, authors, publication year) and unstructured content (abstracts and full texts). These information is  used in 2 ways: generate QA pairs and serve as the RAG retrieval corpus. Our dataset covers 5 major disciplines: physics, chemistry, biology, geography, and mathematics, ensuring broad topical diversity and scientific relevance. All records were imported into a vector database~\cite{qdrant}, enabling efficient dense retrieval. The abstract serves as the primary source for both question generation and retrieval context.

\subsection{Synthetic Multi-Step Reasoning Q-A Pairs}
\label{qa_generate}
To construct QA pairs from scientific literature, we apply a dual-method approach that includes both single-hop and multi-hop question generation techniques. This ensures a diverse set of questions ranging from basic fact extraction to multi-step reasoning.
For single-hop question generation, we adopt the \textit{LMQG}~\cite{lmqg} framework, which uses a T5-based model (\texttt{t5-small-squad-qag}) to jointly generate answer spans and corresponding questions from input passages. Specifically, we use the unsupervised mode of LMQG where the full abstract text is passed to the model, and the pipeline internally extracts candidate answer spans before generating natural language questions conditioned on both the context and the identified answers. This allows for large scale, automatic generation of factoid style questions from scientific abstracts with minimal human intervention.

For multi-hop question generation, we adopt the \textit{Unsupervised-Multi-hop-QA}~\cite{pan-etal-2021-unsupervised} framework to construct questions that require reasoning across two semantically linked contexts. The overall process is illustrated in Figure~\ref{dataset_construction}. Specifically, for each source abstract in our corpus, we first retrieve its most semantically similar abstract based on dense embeddings (BGE embeddings via the Qdrant index~\cite{qdrant}). The resulting pair the original abstract and its nearest neighbor is fed into the multi-hop pipeline, which operates in four steps: (1) shared named entities (bridge entities) between the two abstracts are identified via NER tagging; (2) for each abstract, candidate questions are generated using a T5-based question generator (\texttt{valhalla/t5-base-qg-hl}), retaining only those that mention at least one bridge entity; (3) the retained questions from one passage are re-generated using the answer from the other passage, enabling cross-context answer linkage; (4) the two partial questions are fused by replacing the bridge entity in one question with the descriptive clause from the other, and a BERT-based fill in the blank model is applied to resolve the resulting gaps into fluent multi-hop questions. This setup is designed to simulate real-world scientific reasoning scenarios where answers span multiple papers.

\subsection{Synthetic Reranking-Oriented Q-C-A Pairs}
The Q-C-A pairs reuse the questions and answers generated in Section~\ref{qa_generate} , with candidate contexts constructed under different settings. WFang et al.~\cite{fang2024enhancing} analyzed how different types of noisy retrieved texts affect the performance of LLMs. They identified 3 common noise types: contexts that are superficially related to the query but lack the correct answer,contexts that are irrelevant to the query , and contexts that are topically related to the query but contain incorrect information. Their findings show that these types of context can easily confuse large language models. 
To evaluate the ability of rerankers towards scientific domains, we construct the reranking-oriented Q-C-A pairs in this paper. As illustrated in Figure~\ref{dataset_construction}, we construct 3 types of disruptive contexts as follows:

\textit{\textbf{Noisy Contexts (NC)}}: A common challenge in real-world scientific retrieval scenarios is the inevitable presence of irrelevant or unrelated passages mixed in with relevant contents~\cite{mirza2024large,kwiatkowski2019natural,ling2017program,sun2024scieval}. To simulate this practical challenge and evaluate the ability of rerankers to distinguish relevant information within noisy data, we design the Noisy Contexts setting. Each query is paired with 5 passages confirmed to be relevant (retrieved via the multi-hop construction pipeline in Appendix~\ref{app:related-work}, Algorithm~\ref{alo:dataset}) and 95 randomly sampled unrelated passages from different scientific domains to ensure genuine irrelevance. This setup serves as a fundamental test for assessing the discernability of rerankers towards noisy contexts.

\textit{\textbf{Semantically Similar but Logically Irrelevant Contexts (SSLI)}}: Passages that share high semantic similarity with the query but are logically irrelevant often appear as challenge. Such contexts can easily mislead rerankers~\cite{qiu2025phybench,jin2019pubmedqa}, as they match surface level keywords and phrases, yet do not contain the necessary information to answer the question. This mismatch between semantic similarity and logical relevance requires rerankers to possess strong reasoning and contextual understanding abilities. To simulate this challenge and rigorously evaluate rerankers’ reasoning skills, we design the Semantically Similar but Logically Irrelevant setting. The SSLI passages are retrieved by embedding cosine similarity ($\geq$0.75 using BGE) with the query, sharing domain terminology but confirmed to not contain the answer. Each query is paired with 90 standard candidate passages and 10 SSLI passages. This setup aims to test the model’s capacity for fine grained logical discrimination, going beyond simple semantic matching to accurately identify answer bearing content.

\textit{\textbf{Counterfactual Contexts (CC)}}: Retrieval passages may contain information that is factually incorrect or contradictory to established knowledge~\cite{abdallah2025rankify,qin2025scihorizon,wu2025apbench,singh2022scirepeval}. Such passages pose a significant challenge because they can mislead rerankers relying solely on semantic similarity, requiring models to assess not just relevance but also factual correctness. To simulate this challenge and rigorously evaluate rerankers’ counterfactual reasoning abilities, we design the Counterfactual Contexts setting. The CC passages are generated by LLM-based rewriting of relevant passages, where key factual entities (numbers, names, causal relationships) are systematically replaced with plausible but incorrect alternatives. Each question includes 90 standard candidate passages and 10 counterfactual passages. This setup tests the model’s ability to discern truthfulness and accuracy in addition to semantic alignment. This setting presents a more realistic and stringent scenario for scientific QA systems, where distinguishing between fact and fiction is essential for trustworthy ouputs.

\subsection{Human Evaluation}
\label{sec:human_eval}

To validate the quality of our synthetically constructed datasets, we conduct a human evaluation study across all four task settings. We recruit three annotators with scientific backgrounds to assess each context along five dimensions: (1) \textit{Semantic Similarity} measures topical and conceptual closeness between the context and the question on a 0--1 scale; (2) \textit{Fluency} assesses grammaticality and readability (0--5); (3) \textit{Coherence} examines internal logical consistency within the context (0--5); (4) \textit{Distractiveness} captures the extent to which the context could mislead a model into an incorrect answer (0--5); and (5) \textit{Factual Accuracy} measures the truthfulness of statements relative to established scientific facts (0--5). Different task settings evaluate different subsets of these metrics based on their design objectives, with inapplicable metrics marked as ``--''.

\begin{table}[htbp]
\centering
\small
\caption{Human evaluation results for four task settings. Metrics not applicable to a given task are marked as ``-''.}
\label{tab:human_eval_all}
\resizebox{\linewidth}{!}{%
\begin{tabular}{lccccc}
\toprule
\textbf{Task} & \textbf{Sem.\ Sim.} & \textbf{Fluency} & \textbf{Coherence} & \textbf{Distract.} & \textbf{Fact.\ Acc.} \\
\midrule
NC        & --    & 4.81  & 4.91  & --    & --    \\
SSLI      & 0.83  & 5.00  & 4.77  & 4.34  & 2.08  \\
CC        & --    & 4.94  & 4.38  & 4.75  & 1.17  \\
Multi-Hop & --    & 4.63  & 4.77  & --    & --    \\
\bottomrule
\end{tabular}
}
\end{table}

As shown in Table~\ref{tab:human_eval_all}, all four task settings achieve high Fluency (4.63-4.94) and Coherence (4.38-4.91) scores, confirming that our synthetically constructed passages are linguistically fluent and internally consistent across tasks. The SSLI contexts show strong Semantic Similarity (0.83) to the questions while maintaining low Factual Accuracy (2.08) and high Distractiveness (4.34), demonstrating that they are topically aligned yet factually misleading. For the CC setting, the low Factual Accuracy (1.17) combined with high Distractiveness (4.75) confirms that the counterfactual passages are effective distractors. The Coherence score for CC (4.38) is notably lower than other settings, reflecting the inherent self-contradictions introduced in counterfactual passages. The NC and Multi-Hop contexts, derived from real scientific literature, achieve the highest Fluency and Coherence scores, validating that the retrieval corpus consists of well-written academic content.

\section{Experiment}
\label{Experiment}
\subsection{Experimental Settings}
\paragraph{Rerankers}
We categorize all rerank baselines into 8 groups based on their architectures. (1) \textit{Dense} rerankers: BGE~\cite{chen2024bge}, BCE~\cite{youdao_bcembedding_2023}, 
Jina~\cite{jina} and MXBAI~\cite{rerank2024mxbai}.
(2) \textit{Sparse lexical} rerankers:  SPLADE~\cite{formal2021splade}. 
(3) \textit{Listwise} rerankers:  RankT5~\cite{zhuang2023rankt5} and ListT5~\cite{yoon2024listt5}. 
(4) \textit{Seq2Seq} rerankers: In-Rank~\cite{zhao2023inrank}. 
(5) \textit{Lightweight student} rerankers: Twolar\cite{baldelli2024twolar} and 
MiniLM\cite{wang2020minilm}.
(6) \textit{Late-interaction} rerankers:
ColBert\cite{santhanam2021colbertv2}
(7) \textit{LLM-based} rerankers: RankGPT~\cite{sun2023chatgpt}, LLM2Vec\cite{behnamghader2024llm2vec}
(8) \textit{Agent-based} rerankers:
Rearank~\cite{zhang2025rearank}
\footnote{As the codes of some rerankers, i.e., Rank-R1~\cite{zhuang2025rank} and G-RAG~\cite{dong2024don}, are unavailable, we do not evaluate these models. Besides, the estimated time of evaluating RankGPT would be approximately 1,611 hours. Thus we do not run RankGPT~\cite{sun2023chatgpt} in this paper.}.

\begin{figure*}[htbp]
  \centering
    \includegraphics[width=\linewidth]{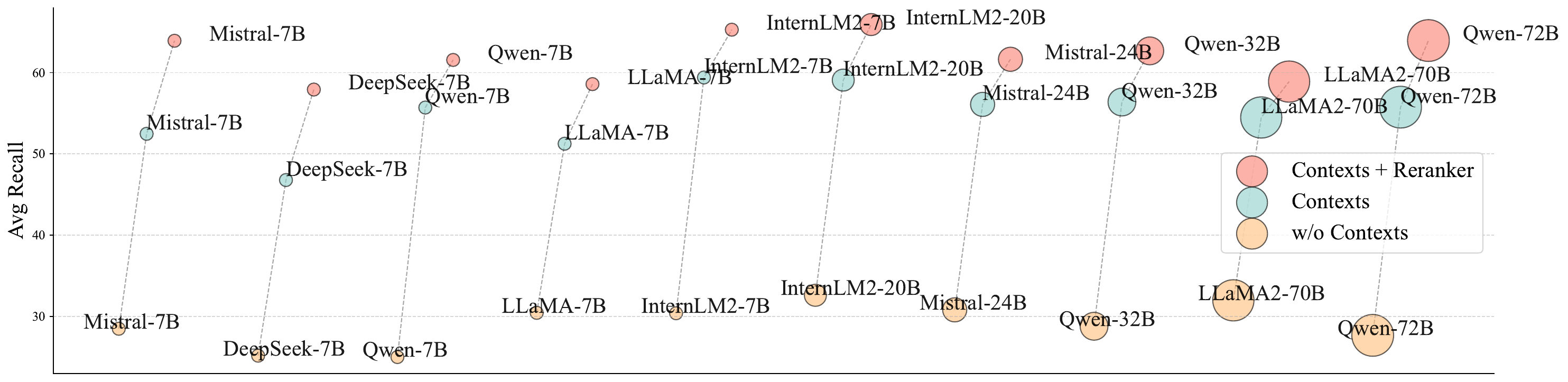}
    \caption{Evaluation results under 3 different settings, averaged across all evaluated rerankers: without any context, with retrieved contexts, and with contexts after applying the reranking stage by recall scores. Error bars represent standard deviation across rerankers.}
    \label{fig:llm-rag}
\end{figure*}

\paragraph{LLMs}
In our experiments, we evaluate a diverse set of 11 open-source LLMs drawn from different families, as summarized in Table~\ref{tab:llm_models} in the Appendix. Mistral models (7B and 24B)~\cite{jiang2023mistral7b} incorporate architectural enhancements such as \textit{Sliding Window Attention (SWA)} and \textit{Grouped-Query Attention (GQA)} to improve long-context modeling and computational efficiency. The LLaMA2-70B model~\cite{touvron2023llama} similarly employs GQA along with \textit{Rotary Positional Embeddings (RoPE)} to better handle extended input sequences. The Qwen model~\cite{qwen2} uses the standard \textit{multi-head attention (MHA)} mechanism, combined with GQA and RoPE, to support multi-language and long-context input while maintaining performance. DeepSeek-V3~\cite{liu2024deepseek} introduces a sparse attention mechanism, \textit{mixed experts (MoE)} and \textit{multi-layer attention (MLA)} to improve large-scale reasoning efficiency and knowledge capacity. We include InternLM2.5~\cite{cai2024internlm2}, a multilingual, open-weight transformer model supporting long-context understanding (up to 200K tokens) and optimized for both instruction-following and reasoning tasks.

\paragraph{Evaluation Protocols and Metrics}
To comprehensively evaluate model performance, we design a two-level evaluation strategy targeting both rerankers and LLMs. For rerankers, we compare the top contexts against annotated relevance labels to assess their ability to retrieve the most informative and factually correct passages for a given scientific question. 
We report \textit{Recall@10}, \textit{Recall@5}, \textit{Average Precision (AP)}, and \textit{Mean Reciprocal Rank (MRR)} to assess ranking quality at different depths.
For LLMs, we assess their ability to generate accurate and complete answers by comparing outputs to golden answers.
We use a token level metrics: \textit{Recall}, which captures how completely the model reproduces the reference content.
This evaluation framework allows us to analyze the individual contribution and limitations of rerankers versus generation models in the RAG pipeline, especially under challenging scientific question answering scenarios. Definitions of these metrics are provided in Appendix~\ref{app:llm-metrics}. 
This two-level evaluation helps us figure out whether mistakes come from retrieving the wrong information, or from the model not using the information well when generating answers. By looking at both parts separately, we can better understand where the system needs improvement. This is especially important for handling complex scientific questions where both accurate retrieval and strong reasoning matter.This approach also helps compare different system components fairly, making it easier to identify which part contributes most to the overall performance.
Moreover, evaluating reranking and generation jointly allows us to examine whether improvements in ranking quality consistently translate into better downstream answers. 
\paragraph{Implementation Details.} The rerankers are conducted on 4 NVIDIA A100 GPUs. All LLMs and rerankers are used with their default configurations. We conduct 3 independent experiments to ensure the reliability and stability of our evaluation results. To ensure a fair comparison, all models are evaluated under a zero-shot setting, without task-specific fine-tuning or additional adaptations.
We collect 100,000 scientific articles and store their abstracts in a Qdrant vector database~\cite{qdrant} to facilitate dense retrieval for RAG-based systems.
Then we preprocess and filter the abstract texts to ensure quality and consistency. Specifically, we retained only those abstracts with a length between 100 and 500 characters. This range was selected to exclude low information or overly terse abstracts, e.g., placeholder texts, while also avoiding excessively long passages that could complicate automatic question generation or overload retrieval models. Following previous works~\cite{yu2024rankrag,abdallah2025rankify,jin2024flashrag}, 100 candidate contexts are first retrieved with BGE as the retriever, then the top 10 are selected by rerankers, which are used as input to LLMs for final answer generation.
In addition, a sampling test dataset showed that RankGPT required 3.5 hours to process 100 samples. Based on this, the estimated time to complete the evaluation on the full dataset would be approximately 1,611 hours. Due to this high computational cost, we do not run RankGPT in this paper.

\begin{figure*}[htbp]
    \centering
    \includegraphics[width=\linewidth]{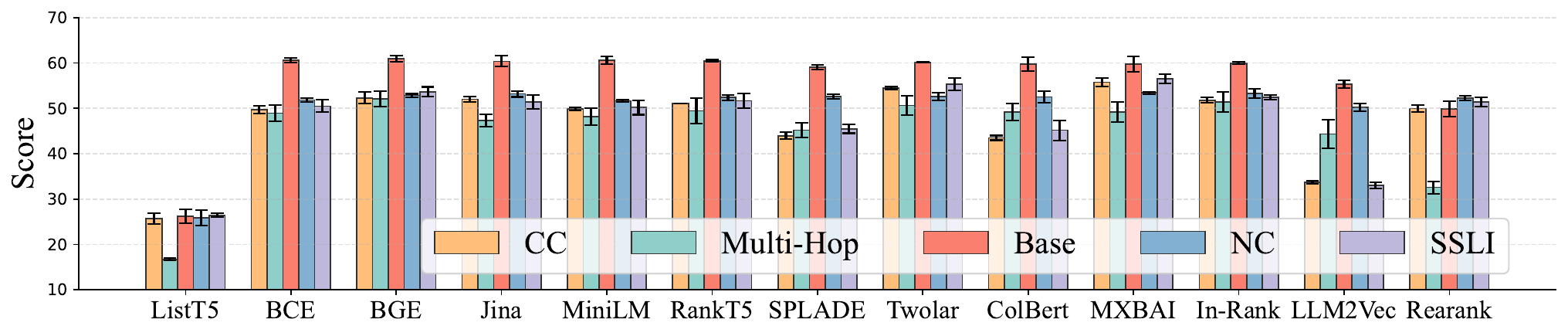}
    \caption{Evaluation of rerankers at \underline{generation stage} on five tasks using \textit{Recall} metric. }
    \label{Generate Recall}
\end{figure*}

\subsection{Evaluating the Usefulness of Rerankers in RAG Pipelines.}

To evaluate the impact of context retrieval and reranking on answer generation quality, we compare LLM performance across 3 distinct settings: (1) zero-context generation (no retrieved context), (2) naive retrieval (Top-10 dense retrieval without reranking), and (3) RAG with reranked contexts (Top-10 after reranking). As shown in Figure~\ref{fig:llm-rag}, all architectures benefit significantly from adding external contexts, with contexts alone contributing 20–30 point gains in recall on average. Among all families, InternLM and Qwen exhibit the strongest improvements from reranking, suggesting that these models better leverage high quality contexts when ranked effectively. InternLM2-20B consistently achieves the highest recall across all settings, while Qwen models also demonstrate robust gains, likely aided by their multilingual pretraining and long-context optimization. In contrast, DeepSeek models show moderate performance without context but respond well to retrieval, albeit with limited further gains from reranking.

\begin{takeaway}[Takeaways]
    \begin{itemize}[leftmargin=1.3em,topsep=1pt,noitemsep]
        \item While all LLMs benefit from retrievers and rerankers, the ability to exploit high quality reranked context varies across different families and architectures.
        \item After reranking the context with rerankers, InternLM and Qwen achieve the best performance.
    \end{itemize}
    
\end{takeaway}

\subsection{Evaluating the Discernability of Rerankers towards Noisy Contexts}

\begin{figure*}[htbp]
    \centering
    \includegraphics[width=\linewidth]{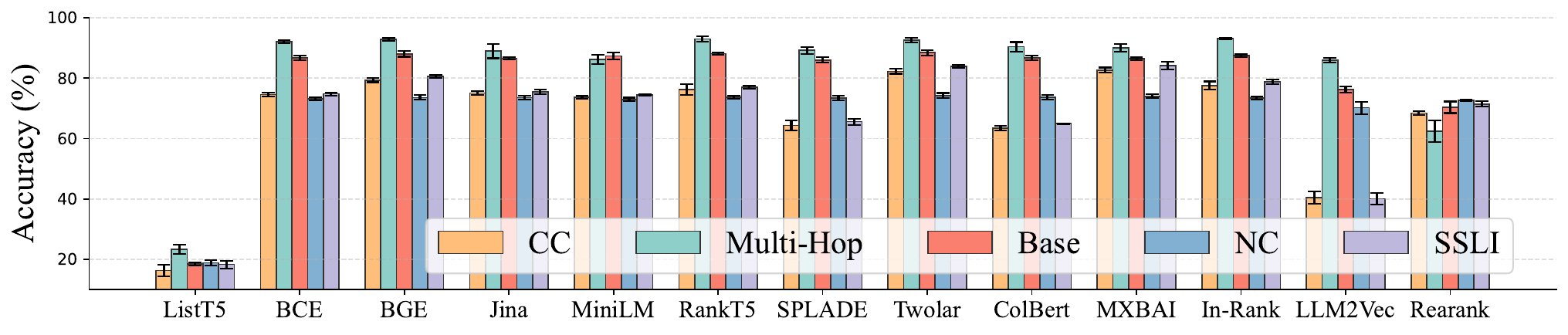}
    \caption{Evaluation of rerankers at \underline{rerank stage} on five tasks using \textit{Recall@10} metric. }
    \label{Rerank Recall}
    \vspace{-3mm}
\end{figure*}

In the \textit{NC} dataset, where 5 relevant contexts are mixed with 95 random distractors. We observe that most rerankers are able to retrieve nearly all relevant contexts within the top-10 results, as indicated by the high recall@10 scores across most evaluated models (Figure~\ref{Rerank Recall}). The presence of a large number of unrelated contexts appears to disperse model attention, impairing the models' ability to maintain coherent focus on relevant information. Even models with strong multi context reasoning capabilities often struggle to prioritize truly important passages over distractors. As a result, the effective recall of useful information during answer generation degrades, leading to lower final answer quality (Figure~\ref{Generate Recall}). Furthermore, when irrelevant contexts are intermixed, models may mistakenly rely on misleading information during both retrieval and generation stages. This phenomenon highlights the necessity of not only retrieving relevant documents but also ensuring high contextual purity. Doing so helps avoid attention diffusion and knowledge contamination in scientific question answering pipelines.

\begin{takeaway}[Takeaways]
    \begin{itemize}[leftmargin=1.3em,topsep=1pt,noitemsep]
        \item LLMs often struggle to effectively filter out irrelevant contexts when handling tasks with a large amount of unrelated information, which affects the quality of final answers.
        \item Effective scientific QA requires both precise retrieval of relevant information and minimizing the influence of irrelevant content.
    \end{itemize}
\end{takeaway}

\subsection{Evaluating the Reasoning Ability of Rerankers towards Semantically Similar but Logical-Irrelevant Contexts}
\label{Counter factual and Semantic Confuse}
The \textit{SSLI} dataset focus on evaluating rerankers' ability to handle semantically challenging cases, where contexts are  semantically similar yet unanswerable. In these complex scenarios, cross-encoder architectures demonstrate significant advantages (Figure~\ref{Rerank Recall}). By jointly encoding the query and document, cross-encoders allow fine-grained feature interactions, capturing subtle semantic differences crucial for distinguishing between true and misleading information. In particular, MXBAI achieves the highest recall scores across nearly all evaluation settings, highlighting the strength of cross-encoder designs in complex reasoning tasks.In contrast, sparse retrieval models, which rely on masked language model objectives to produce sparse representations, exhibit significantly lower recall (approximately 44\%). Although sparse models offer faster inference speed achieving roughly 71.4\% the inference time of dense architectures their limited semantic granularity hinders their ability to identify fine-grained factual inconsistencies, leading to notable performance degradation on nuanced reasoning tasks. Late interaction models such as ColBert, which independently encode queries and documents before performing token wise interaction, maintain strong recall in random distractor settings but struggle when deeper semantic comprehension is required. ColBert’s recall drops to 43.47\% on counterfactual tasks and 45.17\% on semantic confusion tasks, illustrating the limitations of independent representation without full context fusion. LLM-based embedding methods like LLM2Vec underperform on semantically challenging tasks, scoring only 33.72\% on counterfactual and 33.04\% on semantic confusion. Although it leverages frozen LLMs for dense representation, the lack of joint encoding limits its ability to capture subtle semantic mismatches crucial for accurate reranking.

\begin{takeaway}[Takeaways]
    \begin{itemize}[leftmargin=1.3em,topsep=1pt,noitemsep]
        \item Cross-encoders outperform other rerankers on semantically challenging tasks, due to their fine-grained query-document interaction.
    \end{itemize}
\end{takeaway}

\subsection{Evaluating the Ranking Ability of Rerankers}

To evaluate the effectiveness of rerankers independently of final answer generation, we additionally present recall@10 to directly measure how well rerankers prioritize relevant information in the top-ranked results (Figure~\ref{Rerank Recall}).For Recall@10, most rerankers achieve relatively high scores across datasets, including multi-hop and complex QA tasks. However, a discrepancy emerges when examining the final answer quality: despite high Recall@10, the generated answers from LLMs are often of lower quality compared to those from the base retrieval-only setup . This indicates that, while rerankers are effective in ensuring that relevant contexts are included within the top-10 candidates, the ultimate answer quality remains constrained by the LLM’s own generation capabilities. The finding highlights a limitation where retrieval improvements alone cannot fully compensate for deficiencies in the LLM’s ability to synthesize and reason over retrieved knowledge. In multi-hop tasks, rerankers can successfully retrieve sufficient relevant context. However, even with ample context, LLMs fail to achieve expected performance due to inherent limitations in their reasoning capabilities, which ultimately restrict the quality of final answers.
Beyond limitations in LLM generation, differences in reranker design also contribute to performance gaps across tasks. Although Rearank uses a listwise reinforcement learning framework with local window‑based ranking over sliding windows, its architecture lacks global cross-document encoding or token-level interaction. As a result, it cannot effectively aggregate complementary evidence scattered across passages, leading to lower Recall@10 (62.41) on multi-hop reasoning tasks.

\begin{takeaway}[Takeaways]
    \begin{itemize}[leftmargin=1.3em,topsep=1pt,noitemsep]
        \item For multi-hop tasks, retrieval and rerankers are able to select high-relevant contexts, but the final performance still depends on the inherent reasoning capacity of LLMs.
        \item Reranker architectures also influence performance on muti-hop task.
    \end{itemize}
\end{takeaway}

\begin{figure}[htbp]
  \centering
  
    \includegraphics[width=0.9\linewidth]{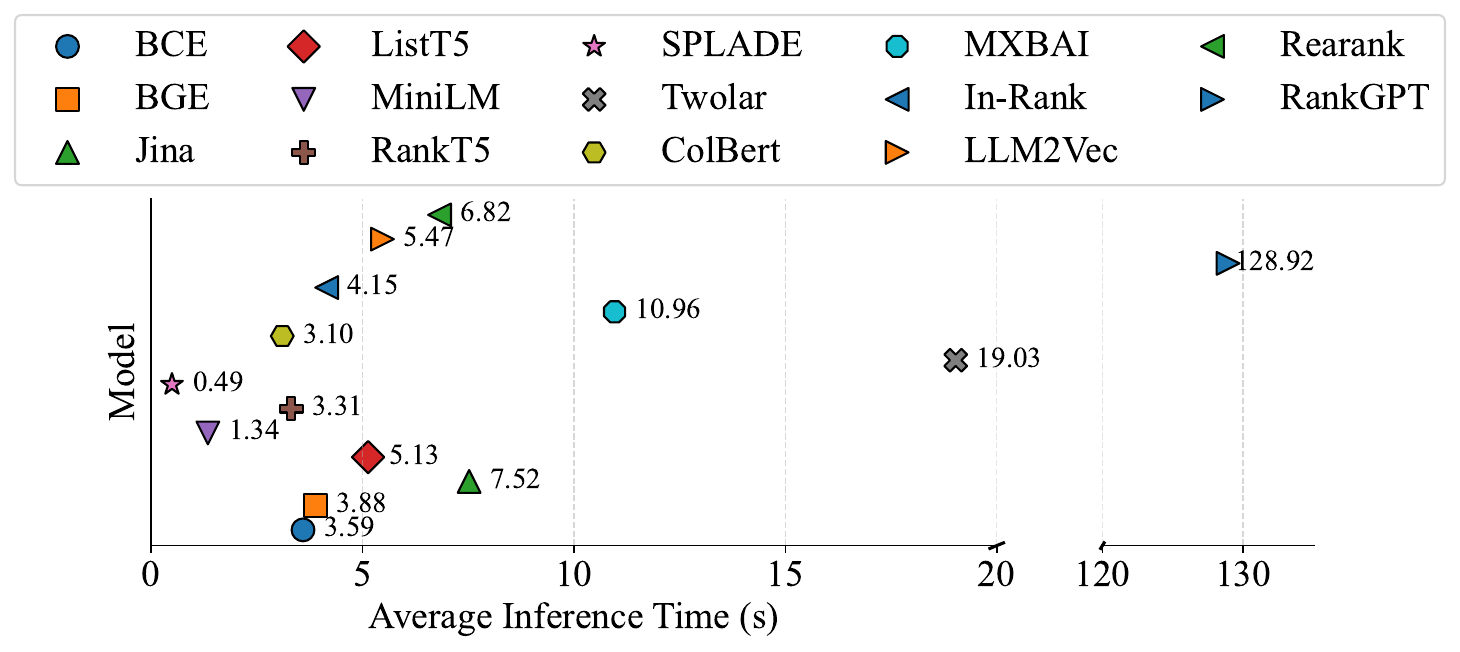}
    \caption{Running efficiency analysis of rerankers.}
   \label{fig:infer_time}
   \vspace{-3mm}
\end{figure}

\subsection{Evaluating the Running Efficiency of Rerankers}
\label{rerank time}
This section analyze the running efficiency. The average inference time of one question is shown in Figure~\ref{fig:infer_time}. We observe clear running efficiency differences across reranker architectures. Sparse models like SPLADE achieve the fastest inference (0.49s), owing to their non-interactive, inverted-index design. Lightweight sentence encoders such as MiniLM also offer rapid performance (1.34s), making them attractive for latency-sensitive applications. In contrast, cross-encoder models demonstrate a trade-off: 
BCE and BGE are relatively efficient (3.5s), while deeper or more expressive models like MXBAI incur higher cost (10.96s) due to full query-document interaction. 
Twolar emphasizes ranking accuracy through a two-stage, listwise-aware architecture. Thus, it takes more time than others. 
LLM2Vec performs dense vector similarity using frozen large language models. Its inference time (5.47s) reflects the cost of extracting and comparing hidden representations from a large LLM, despite not involving decoding. 
Rearank leverages an agent-based framework with iterative reasoning steps guided by reinforcement learning. Its higher latency (6.82s) is mainly attributed to the multi-round LLM reasoning process required to evaluate and reorder candidate documents.
RankGPT leverages the reasoning power of large language models for reranking through a generative, prompt-based approach. Its extremely slow inference speed, which is caused by having to process each candidate document one at a time through a LLM.

\begin{takeaway}[Takeaways]
    \begin{itemize}[leftmargin=1.3em,topsep=1pt,noitemsep]
        \item Better reranking performance often comes at the cost of increased inference time.
    \end{itemize}
\end{takeaway}

\begin{figure}[htbp]
    \centering
    \includegraphics[width=\linewidth]{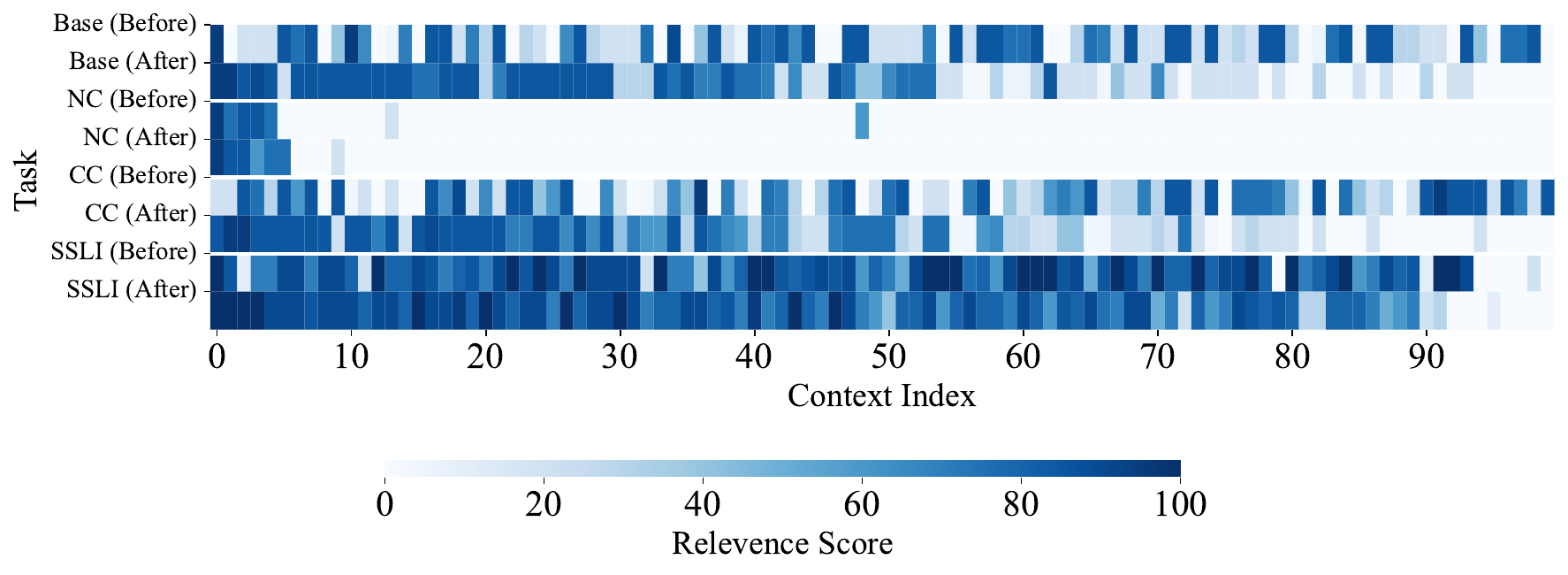}
    \vspace{-4mm}
    \caption{Visualization of question-context relevance (before and after rerankering) under 4 evaluation tasks.}
    \vspace{-4mm}
    \label{fig:heapmap}
\end{figure}

\subsection{Question-Context Relevance Under Four Task Settings}
To better understand the effectiveness of the rerankers in distinguishing relevant from irrelevant contexts, we visualize the context ranking distributions before and after reranking in 4 tasks.
We use DeepSeek-V3 671B to assess the semantic relevance between contexts and the given question.
The visualization results of the relevance heatmap are shown in Figure~\ref{fig:heapmap}. In the heatmap visualization, we group the results of the same task into pairs, allowing us to observe changes in color intensity to evaluate the effectiveness of the reranking.
An overview of the 4 tasks reveals that after reranking, the relevant contexts are clearly moved to the top positions across all tasks. Particularly in the NC task, even when faced with a large amount of irrelevant context, the rerankers is still able to correctly prioritize the relevant contexts. However, in the SSLI and CC tasks, the model incorrectly ranks irrelevant contexts among the top , leading to poor performance in LLM as discussed in Section ~\ref{Counter factual and Semantic Confuse}. This suggests that the rerankers limitations in its ability to perform fine-grained semantic reasoning beyond surface-level relevance.

\begin{takeaway}[Takeaways]
\begin{itemize}[leftmargin=1.3em,topsep=1pt,noitemsep]
    \item Rerankers effectively promote relevant contexts to higher positions across multiple tasks, especially in noisy settings, but still struggle in scenarios that require deep semantic understanding beyond surface-level similarity.
\end{itemize}
\end{takeaway}


\section{Conclusion}
In this paper, we introduce SciRerankBench. To the best of our knowledge, it is the first publicly available benchmark for evaluating rerankers in RAG-LLMs towards scientific domains. We design datasets covering noisy contexts, semantically similar but logically irrelevant contexts and counterfactual contexts comprehensively evaluate the rerankers. Evaluating 13 rerankers and 11 LLMs on these tasks, our experiments show that while rerankers struggle to filter logically irrelevant yet semantically similar passages. Moreover, final answer quality remains constrained by the inherent reasoning limitations of current LLMs.
These findings highlight the need for logic-aware rerankers and capable LLMs to improve scientific QA.
We hope this benchmark will inspire promising valuable future research work on more robust and reasoning-aware reranking systems.

\section*{Acknowledgements}
The work of Hengshu Zhu was supported by the Strategic Priority Research Program of Chinese Academy of Sciences (No.XDB1350102). The work of Yuanchun Zhou was supported by the National Natural Science Foundation of China (No. 92470204). The work of Meng Xiao were supported by the National Natural Science Foundation of China (No. 62506351) and Beijing Natural Science Foundation (No.4254089).

\section*{Limitations}
\label{Limitations}
This benchmark focuses exclusively on open-source LLMs. Closed-source models such as GPT-4 or Claude are not included due to limited access, which may restrict the generality of our observations to fully proprietary systems.

\section*{Ethics Statement}
\label{sec:ethics}
This research uses only publicly available scientific literature metadata from OpenAlex~\cite{priem2022openalex}, which is released under open licenses.  All models and datasets evaluated in this work are publicly released, and our benchmark code and data are made openly available to promote reproducibility and fairness in the scientific community.

\bibliography{cite}

\clearpage
\appendix
\section{Appendix}

\subsection{Related Work (Extended)}
\label{app:related-work}

\paragraph{Scientific Question Answering Datasets}

Existing QA datasets fall into two categories: open-domain and scientific. Open-domain datasets include Natural Questions (NQ)~\cite{kwiatkowski2019natural}, TriviaQA~\cite{joshi2017triviaqa}, and WebQuestions~\cite{wang2014overview}, while HotpotQA~\cite{yang2018hotpotqa} and MuSiQue~\cite{trivedi2022musique} emphasize multi-hop reasoning. In scientific domains, BioASQ~\cite{tsatsaronis2012bioasq} and PubMedQA~\cite{jin2019pubmedqa} target biomedical literature, while MATH~\cite{hendrycks2021measuring} and AquaRat~\cite{ling2017program} evaluate mathematical reasoning. General science benchmarks like OpenBookQA~\cite{mihaylov2018can} and ARC~\cite{clark2018think} combine retrieval with multi-step reasoning.

Prior work~\cite{abdallah2025rankify,jin2024flashrag} shows that question-related contexts help LLMs generate more reliable answers. However, except for PubMedQA and BioASQ, most datasets do not provide rich contextual information and assume relevant context is already retrieved. They fail to simulate real-world conditions where models must reason over noisy or misleading documents, limiting their ability to evaluate rerankers.

\begin{algorithm}[h]
\caption{Dataset Construction Process}
\label{alo:dataset}
\KwIn{Source scientific corpus $C$ }
\KwOut{QA Pairs $D_{QA}$, Reranking Set $D_{rank}$}

\textbf{// Step 1: Preprocessing \& Indexing}\;
$C' \leftarrow \text{Filter abstracts: } \text{len} \in [100, 500]$\;
$V \leftarrow E_{bge}(C')$\;
Insert($V$ into Qdrant index $I$)\;

\vspace{1mm}
\textbf{// Step 2: QA Generation}\;
\ForEach{$a \in C'$}{
    $(q_1, ans_1) \leftarrow \text{LMQG}(a)$\;
    Add $(q_1, ans_1)$ to $D_{QA}^{single}$\;

    $a' \leftarrow \text{NearestNeighbor}(a, I)$\;
    \If{$a' \neq \text{NULL}$}{
        $(q_2, ans_2) \leftarrow \text{UMQG}(a, a')$\;
        Add $(q_2, ans_2)$ to $D_{QA}^{multi}$\;
    }
}

\vspace{1mm}
\textbf{// Step 3: Construct Reranking Subsets}\;
\ForEach{$(q, ans) \in D_{QA}^{single}$}{
    $P \leftarrow \text{DenseRetrieve}(q, I, k=100)$\;

    $P_{rand} \leftarrow \text{Mix}(5 \cdot \text{Rel}, 95 \cdot \text{Random})$\;
    $P_{conf} \leftarrow \text{InjectConfusers}(P, 10)$\;
    $P_{cfact} \leftarrow \text{InjectCounterfacts}(P, 10)$\;

    Add $(q, P_{rand}, ans)$ to $D_{rank}$\;
    Add $(q, P_{conf}, ans)$ to $D_{rank}$\;
    Add $(q, P_{cfact}, ans)$ to $D_{rank}$\;
}

\vspace{1mm}
\textbf{// Step 4: Final Evaluation Set}\;
$D_{\text{eval}} \leftarrow D_{rank} \cup D_{QA}^{multi}$\;

\end{algorithm}


\paragraph{Scientific Question Answering Benchmarks}

Scientific QA benchmarks are widely used to evaluate LLMs. Recent benchmarks target specific domains, such as APBench~\cite{wu2025apbench}, ChemBench~\cite{mirza2024large}, and PHYBench~\cite{qiu2025phybench}, but offer limited cross-domain generalizability. Multidisciplinary benchmarks including SciEval~\cite{sun2024scieval}, SciBench~\cite{wang2023scibench}, AGIEval~\cite{zhong2023agieval}, SciRepEval~\cite{singh2022scirepeval}, and SciHorizon~\cite{qin2025scihorizon} evaluate LLMs across diverse domains and difficulty levels. Rankify~\cite{abdallah2025rankify} and FlashRAG~\cite{jin2024flashrag} provide modular infrastructures for RAG experimentation. However, existing benchmarks primarily assess end-to-end QA without isolating reranking components, while Rankify and FlashRAG use open-domain rather than scientific corpora.

\paragraph{Rerankers in Retrieval-Augmented Generated LLMs}
Reranker reorders retrieved candidate documents based on deeper semantic relevance, improving the accuracy and quality of the final retrieval results.
Rerankers span several architectures. \textit{Dense cross-encoders} (BGE~\cite{chen2024bge}, Jina~\cite{jina}, BCE~\cite{youdao_bcembedding_2023}, MXBAI~\cite{rerank2024mxbai}) compute relevance scores via bi-sequence encoders. \textit{Sparse lexical} models like SPLADE~\cite{formal2021splade} use term-level matching with sparsity constraints. Lightweight alternatives like MiniLM~\cite{wang2020minilm} reduce inference cost via distillation. \textit{Late-interaction} models like ColBERT~\cite{santhanam2021colbertv2} delay token-level interaction to scoring. \textit{LLM-based} rerankers (RankGPT~\cite{sun2023chatgpt}, LLM2Vec~\cite{behnamghader2024llm2vec}) leverage LLM reasoning at high computational cost. \textit{Seq2Seq} rerankers (In-Rank~\cite{zhao2023inrank}) use generative decoding for relevance assessment. \textit{Listwise} models (RankT5~\cite{zhuang2023rankt5}, ListT5~\cite{yoon2024listt5}) optimize global candidate ordering. Twolar~\cite{baldelli2024twolar} transfers LLM knowledge to lightweight students. Rank-R1~\cite{zhuang2025rank} uses RL with answer-aware rewards. G-RAG~\cite{dong2024don} models document connections via GNNs. \textit{Agent-based} rerankers like Rearank~\cite{zhang2025rearank} iteratively examine passages via RL.

However, existing benchmarks do not independently evaluate these rerankers. They typically assess end-to-end RAG systems assuming high-quality retrieved contexts, making it difficult to assess reranker ability to distinguish relevant from misleading passages in scientific tasks.

\subsection{Detailed Dataset Construction Information}
\paragraph{Source Literature}
We choose OpenAlex~\cite{priem2022openalex} as our source data to construct our synthetic scientific questions. OpenAlex is an open, comprehensive scholarly metadata platform that provides structured information on research papers, authors, institutions, concepts, and more. 
This database contains over 250 million works, more than 100 million unique author profiles, and metadata for approximately 120,000 publication venues, including both journals and conferences.


\paragraph{Dataset Construction Pseudocode} Algorithm~\ref{alo:dataset} outlines the dataset construction pipeline, including abstract preprocessing, single- and multi-hop QA pair generation, and the creation of three datasets which designed to evaluate different challenges.

\paragraph{Specific Dataset Instances/Samples}
Here we provide a specific dataset instances or sample, which comprising a scientific question, its corresponding golden answer, and a set of 100 contextual passages. The sample is shown in the Figure~\ref{fig:dataste-sample-1} to Figure~\ref{fig:dataste-sample-3}.

\paragraph{Dataset Statistics.}

Table~\ref{tab:qa_counts} provides a detailed breakdown of the number of QA instances across five scientific subjects Biology, Math, Physics, Geology, and Chemistry under five distinct task types. These task types correspond to the benchmark datasets described previously: NC,Base,CC,Multi-hop,SSLI.

\begin{table}[ht]
\centering
\small
\caption{QA Counts for different subjects and tasks.}
\begin{tabular}{ccc}
\toprule
\textbf{Subject} & \textbf{Task Type} & \textbf{Number} \\
\midrule
\multirow{5}{*}{Biology} & NC & 2499 \\
                         & Base & 2499 \\
                         & CC  & 2496 \\
                         & Multi-Hop    & 1246 \\
                         & SSLI  & 2497 \\
\midrule
\multirow{5}{*}{Math}    & NC  & 2494 \\
                         & Base & 2494 \\
                         & CC  & 2491 \\
                         & Multi-Hop    & 1631 \\
                         & SSLI  & 2493 \\
\midrule
\multirow{5}{*}{Physics} & NC  & 2491 \\
                         & Base & 2491 \\
                         & CC  & 2494 \\
                         & Multi-Hop    & 1425 \\
                         & SSLI  & 2492 \\
\midrule
\multirow{5}{*}{Geology} & NC  & 2493 \\
                         & Base & 2493 \\
                         & CC  & 2493 \\
                         & Multi-Hop    & 1598 \\
                         & SSLI  & 2496 \\
\midrule
\multirow{5}{*}{Chemistry} & NC & 2499 \\
                           & Base & 2499 \\
                           & CC  & 2497 \\
                           & Multi-Hop  & 1087 \\
                           & SSLI  & 2498 \\
\bottomrule
\end{tabular}

\label{tab:qa_counts}
\end{table}

\begin{table*}[htbp]
\centering
\small
\caption{Performance of  LLM-based and Agent-based rerankers across various evaluation tasks (Recall)}
\label{tab:reranker_tasks_LLM_and_Agent_Recall}
\begin{tabular}{ccccccc}
\toprule
\textbf{Reranker} & \textbf{Subject} & \textbf{Multi-Hop} & \textbf{NC} & \textbf{CC} & \textbf{SSLI} & \textbf{Base}  \\
\midrule
\multirow{5}{*}{\textbf{LLM2Vec}} 
& Bio.  & 33.72$\pm$0.36 & 32.77$\pm$0.45 & 31.78$\pm$0.52 & 31.34$\pm$0.61 & 31.46$\pm$0.50 \\
& Geo.  & 44.34$\pm$3.09 & 41.84$\pm$2.70 & 38.45$\pm$2.10 & 37.06$\pm$1.85 & 35.99$\pm$1.75 \\
& Chem. & 55.28$\pm$0.84 & 53.75$\pm$0.88 & 51.35$\pm$1.10 & 49.39$\pm$1.05 & 49.45$\pm$0.90 \\
& Phy.  & 50.18$\pm$0.83 & 47.82$\pm$0.75 & 44.42$\pm$0.82 & 43.96$\pm$0.65 & 45.69$\pm$0.88 \\
& Math. & 33.04$\pm$0.66 & 29.49$\pm$1.10 & 31.05$\pm$0.92 & 29.26$\pm$1.15 & 28.99$\pm$1.90 \\
\midrule
\multirow{5}{*}{\textbf{Rearank}} 
& Bio.  & 49.98$\pm$0.79 & 48.57$\pm$1.27 & 47.11$\pm$0.81 & 46.48$\pm$2.16 & 46.69$\pm$0.94 \\
& Geo.  & 32.56$\pm$1.36 & 30.73$\pm$1.92 & 26.71$\pm$2.38 & 24.83$\pm$0.84 & 23.48$\pm$1.17 \\
& Chem. & 49.87$\pm$1.67 & 48.51$\pm$0.88 & 46.32$\pm$1.21 & 44.56$\pm$1.18 & 44.64$\pm$0.61 \\
& Phy.  & 52.25$\pm$0.46 & 49.76$\pm$0.23 & 46.27$\pm$0.70 & 45.86$\pm$0.32 & 47.64$\pm$0.87 \\
& Math. & 51.45$\pm$1.02 & 45.90$\pm$1.34 & 48.35$\pm$0.50 & 45.62$\pm$1.23 & 45.22$\pm$3.21 \\
\bottomrule

\end{tabular}
\end{table*}

\subsection{Detailed Experimental Settings}
\paragraph{Detailed LLM Information}
Detailed information of LLM models evaluated in this paper is shown in Table \ref{tab:appendix_llm}. These models are from 5 LLM families, i.e., Mistral, LLaMA, DeepSeek, Qwen and InternLM2 , with parameter sizes ranging from 7B to 72B+. This diverse set of LLMs also enables us to introduce a new evaluation perspective using the quality of LLM-generated answers as an indicator for how well a reranker supports real task performance.

\begin{table}[htbp]
\centering
\small
\caption{LLMs evaluated in this paper.}
\begin{tabular}{llll}
\toprule
\textbf{Model} & \textbf{Family} & \textbf{Size} \\
\midrule
Mistral-7B-v0.2 & Mistral & 7B \\
Mistral-24B & Mistral & 24B \\
\midrule
LLaMA-7B & Meta (LLaMA) & 7B  \\
LLaMA2-70B & Meta (LLaMA2) & 70B  \\
\midrule
DeepSeek-V3-7B & DeepSeek & 7B  \\
DeepSeek-V3-671B & DeepSeek & 671B  \\
\midrule
Qwen-7B & Qwen & 7B  \\
Qwen-32B & Qwen & 32B  \\
Qwen-72B & Qwen & 72B  \\
\midrule
InternLM2-7B & InternLM2 & 7B  \\
InternLM2-20B & InternLM2 & 20B  \\

\bottomrule
\label{tab:appendix_llm}

\end{tabular}

\label{tab:llm_models}
\end{table}

\paragraph{Detailed Reranker Information} Detailed information of the rerankers is shown in Table \ref{tab:reranker_models}.

\begin{table}[htbp]
\centering
\small
\caption{RAG reranking methods evaluated in this paper.}
\begin{tabular}{
l
p{2cm}
p{2.8cm}
}
\toprule
\textbf{Model} & \textbf{Category} & \textbf{Reranking Strategy} \\
\midrule
BGE & Transformer & Cross-Encoder \\
Jina & Transformer & Cross-Encoder \\
BCE & Transformer & Cross-Encoder \\
MXBAI & Transformer & Cross-Encoder \\
\midrule
MiniLM & Sentence Transformer  & distilled self-attention \\
\midrule
ColBert & BERT & late interaction mechanism \\
\midrule
In-Rank & Seq2Seq model & tokenizes query-document pairs \\
\midrule
SPLADE & Sparse & MLMs \\
\midrule
RankT5 & T5-based & ranking losses \\
ListT5 & T5-based & m-ary tournament sort \\
Twolar & T5-based & two-step distillation approach \\
\midrule
LLM2Vec & LLM-based & LLM Embedding Similarity \\
RankGPT & LLM-based & sliding window \\
\midrule
Rearank & Agent-based &  reinforcement learning\\
\bottomrule
\end{tabular}

\label{tab:reranker_models}
\end{table}

\begin{table*}[t]
\centering
\small
\caption{Performance of LLM-based and Agent-based rerankers across various evaluation tasks (Recall@10).}
\label{tab:reranker_tasks_LLM_and_Agent_Recall@10}

\begin{tabular}{ccccccc}
\toprule
\textbf{Reranker} & \textbf{Subject} & \textbf{Multi-Hop} & \textbf{NC} & \textbf{CC} & \textbf{SSLI} & \textbf{Base} \\
\midrule

\multirow{5}{*}{\textbf{LLM2Vec}}
& Bio.  & 85.87$\pm$0.82 & 70.18$\pm$2.05 & 40.39$\pm$2.09 & 40.09$\pm$1.89 & 76.20$\pm$1.09 \\
& Geo.  & 83.80$\pm$0.75 & 68.00$\pm$1.85 & 38.50$\pm$1.90 & 38.20$\pm$1.60 & 74.00$\pm$1.10 \\
& Chem. & 84.90$\pm$0.70 & 69.05$\pm$1.95 & 39.20$\pm$1.85 & 39.10$\pm$1.70 & 75.10$\pm$1.05 \\
& Phy.  & 81.60$\pm$0.85 & 66.50$\pm$1.50 & 37.50$\pm$1.60 & 37.40$\pm$1.45 & 72.20$\pm$1.25 \\
& Math. & 82.70$\pm$0.80 & 67.20$\pm$1.65 & 37.90$\pm$1.70 & 37.80$\pm$1.50 & 73.10$\pm$1.20 \\
\midrule

\multirow{5}{*}{\textbf{Rearank}}
& Bio.  & 70.35$\pm$0.43 & 72.68$\pm$0.31 & 68.42$\pm$0.64 & 71.43$\pm$0.85 & 70.33$\pm$1.92 \\
& Geo.  & 59.87$\pm$1.70 & 69.67$\pm$1.18 & 62.41$\pm$3.61 & 69.69$\pm$2.24 & 68.03$\pm$1.47 \\
& Chem. & 69.20$\pm$1.31 & 70.18$\pm$1.40 & 68.72$\pm$0.77 & 67.67$\pm$1.38 & 68.83$\pm$1.49 \\
& Phy.  & 60.96$\pm$0.85 & 70.44$\pm$0.44 & 60.96$\pm$2.49 & 66.72$\pm$0.37 & 67.37$\pm$0.69 \\
& Math. & 59.95$\pm$0.72 & 67.77$\pm$0.75 & 61.85$\pm$2.11 & 67.62$\pm$0.65 & 67.17$\pm$1.32 \\
\bottomrule

\end{tabular}
\end{table*}

\begin{table*}[htbp]
\centering
\small
\caption{Performance of LLM-based and Agent-based rerankers across various evaluation tasks (Recall@5).}
\label{tab:reranker_tasks_qwen72b_Recall@5_LLM2Vec_Rearank}
\begin{tabular}{ccccccc}
\toprule
\textbf{Reranker} & \textbf{Subject} & \textbf{Multi-Hop} & \textbf{NC} & \textbf{CC} & \textbf{SSLI} & \textbf{Base} \\
\midrule
\multirow{5}{*}{\textbf{LLM2Vec}}
& Bio.  & 68.49$\pm$7.41 & 66.67$\pm$6.55 & 21.43$\pm$4.21 & 15.78$\pm$3.28 & 64.62$\pm$6.21 \\
& Geo.  & 69.29$\pm$7.41 & 63.80$\pm$6.55 & 20.78$\pm$4.21 & 14.93$\pm$3.28 & 59.84$\pm$6.21 \\
& Chem.  & 66.66$\pm$7.41 & 62.82$\pm$6.55 & 20.26$\pm$4.21 & 14.63$\pm$3.28 & 59.99$\pm$6.21 \\
& Phy.  & 66.62$\pm$7.41 & 63.31$\pm$6.55 & 20.38$\pm$4.21 & 14.78$\pm$3.28 & 61.17$\pm$6.21 \\
& Math. & 67.56$\pm$7.41 & 62.62$\pm$6.55 & 19.92$\pm$4.21 & 14.52$\pm$3.28 & 62.00$\pm$6.21 \\
\midrule
\multirow{5}{*}{\textbf{Rearank}}
& Bio.  & 38.65$\pm$3.09 & 63.26$\pm$0.96 & 54.85$\pm$0.67 & 55.01$\pm$0.84 & 51.81$\pm$2.01 \\
& Geo.  & 36.79$\pm$0.45 & 62.45$\pm$0.89 & 55.49$\pm$0.78 & 54.75$\pm$1.03 & 52.76$\pm$1.32 \\
& Chem.  & 36.51$\pm$1.92 & 61.11$\pm$1.15 & 53.20$\pm$1.58 & 53.29$\pm$1.53 & 52.44$\pm$0.65 \\
& Phy.  & 36.51$\pm$3.81 & 61.41$\pm$0.73 & 53.92$\pm$1.34 & 53.49$\pm$0.37 & 51.37$\pm$1.61 \\
& Math. & 37.28$\pm$1.81 & 60.39$\pm$0.93 & 54.07$\pm$2.63 & 53.14$\pm$1.36 & 53.28$\pm$2.30 \\
\bottomrule
\end{tabular}
\end{table*}

\begin{table*}[htbp]
\centering
\small
\caption{Performance of LLM-based and Agent-based rerankers across various evaluation tasks (AP).}
\label{tab:reranker_tasks_qwen72b_AP_LLM2Vec_Rearank}
\begin{tabular}{ccccccc}
\toprule
\textbf{Reranker} & \textbf{Subject} & \textbf{Multi-Hop} & \textbf{NC} & \textbf{CC} & \textbf{SSLI} & \textbf{Base} \\
\midrule
\multirow{5}{*}{\textbf{LLM2Vec}}
& Bio.  & 51.60$\pm$7.85 & 55.55$\pm$5.61 & 18.85$\pm$3.36 & 14.83$\pm$1.91 & 58.55$\pm$5.56 \\
& Geo.  & 50.86$\pm$7.85 & 54.03$\pm$5.61 & 18.14$\pm$3.36 & 13.94$\pm$1.91 & 53.93$\pm$5.56 \\
& Chem.  & 49.96$\pm$7.85 & 53.27$\pm$5.61 & 18.32$\pm$3.36 & 13.94$\pm$1.91 & 55.07$\pm$5.56 \\
& Phy.  & 50.22$\pm$7.85 & 53.71$\pm$5.61 & 18.71$\pm$3.36 & 14.13$\pm$1.91 & 55.37$\pm$5.56 \\
& Math. & 51.10$\pm$7.85 & 53.46$\pm$5.61 & 18.12$\pm$3.36 & 13.80$\pm$1.91 & 56.48$\pm$5.56 \\
\midrule
\multirow{5}{*}{\textbf{Rearank}}
& Bio.  & 36.57$\pm$2.69 & 48.28$\pm$1.07 & 51.16$\pm$0.62 & 51.82$\pm$0.82 & 48.91$\pm$2.01 \\
& Geo.  & 33.63$\pm$0.87 & 51.93$\pm$1.53 & 53.25$\pm$1.26 & 52.51$\pm$0.90 & 50.58$\pm$1.49 \\
& Chem.  & 34.01$\pm$2.59 & 47.14$\pm$1.73 & 49.96$\pm$1.35 & 50.05$\pm$1.15 & 49.60$\pm$1.32 \\
& Phy.  & 34.64$\pm$3.66 & 46.78$\pm$0.37 & 50.75$\pm$1.24 & 50.47$\pm$0.10 & 49.01$\pm$1.50 \\
& Math. & 35.13$\pm$1.60 & 48.32$\pm$0.80 & 51.53$\pm$2.30 & 50.47$\pm$1.41 & 51.02$\pm$2.37 \\
\bottomrule
\end{tabular}
\end{table*}

\begin{table*}[htbp]
\centering
\small
\caption{Performance of LLM-based and Agent-based rerankers across various evaluation tasks (MRR).}
\label{tab:reranker_tasks_qwen72b_MRR_LLM2Vec_Rearank}
\begin{tabular}{ccccccc}
\toprule
\textbf{Reranker} & \textbf{Subject} & \textbf{Multi-Hop} & \textbf{NC} & \textbf{CC} & \textbf{SSLI} & \textbf{Base} \\
\midrule
\multirow{5}{*}{\textbf{LLM2Vec}}
& Bio.  & 0.52$\pm$0.08 & 0.55$\pm$0.06 & 0.19$\pm$0.05 & 0.14$\pm$0.01 & 0.58$\pm$0.06 \\
& Geo.  & 0.51$\pm$0.08 & 0.54$\pm$0.06 & 0.18$\pm$0.05 & 0.13$\pm$0.01 & 0.54$\pm$0.06 \\
& Chem.  & 0.49$\pm$0.08 & 0.53$\pm$0.06 & 0.18$\pm$0.05 & 0.13$\pm$0.01 & 0.54$\pm$0.06 \\
& Phy.  & 0.50$\pm$0.08 & 0.53$\pm$0.06 & 0.18$\pm$0.05 & 0.13$\pm$0.01 & 0.55$\pm$0.06 \\
& Math. & 0.51$\pm$0.08 & 0.53$\pm$0.06 & 0.18$\pm$0.05 & 0.13$\pm$0.01 & 0.56$\pm$0.06 \\
\midrule
\multirow{5}{*}{\textbf{Rearank}}
& Bio.  & 0.38$\pm$0.03 & 0.54$\pm$0.01 & 0.52$\pm$0.01 & 0.53$\pm$0.01 & 0.50$\pm$0.02 \\
& Geo.  & 0.35$\pm$0.01 & 0.55$\pm$0.02 & 0.54$\pm$0.01 & 0.54$\pm$0.01 & 0.52$\pm$0.02 \\
& Chem.  & 0.35$\pm$0.03 & 0.54$\pm$0.02 & 0.51$\pm$0.02 & 0.51$\pm$0.01 & 0.51$\pm$0.01 \\
& Phy.  & 0.36$\pm$0.04 & 0.54$\pm$0.00 & 0.52$\pm$0.01 & 0.51$\pm$0.00 & 0.50$\pm$0.01 \\
& Math. & 0.37$\pm$0.01 & 0.54$\pm$0.01 & 0.52$\pm$0.02 & 0.52$\pm$0.01 & 0.52$\pm$0.02 \\
\bottomrule
\end{tabular}
\end{table*}

\paragraph{Prompting Template}
To standardize the evaluation of reranking models, we adopt a unified prompting template that guides models to generate concise answers based solely on the provided contexts. As illustrated in Figure~\ref{fig:prompt-template}, the prompt explicitly instructs the model to refrain from using external knowledge and to return ``No Answer Present" when no relevant information is found.

\definecolor{lightgray}{gray}{0.95}
\begin{figure}[htbp]
\centering
\small
\begin{tcolorbox}[colback=gray!5!white, colframe=gray!80!black, title=Our Prompting Template]
\small
\textbf{Instructions:}

You are given a question and a set of contexts. Your task is to provide a clear and concise answer to the question based on the contexts provided.

Answer based \textbf{only} on the contexts. If none are relevant, say: \textbf{No Answer Present.}

Keep your answer short ideally within 2 sentences. Do not include references or restate the question.

\textbf{---}

QUESTION: \{QUESTION\} \\
CONTEXTS: \{CONTEXTS\}

\textbf{\#\#\#} \\
ANSWER:
\end{tcolorbox}
\caption{Prompt used in the Rerank Evaluation Task}
\label{fig:prompt-template}
\end{figure}

\paragraph{Detailed Information of Evaluation Protocols and Metrics}
\label{app:llm-metrics}
For rerankers, we compare the top-ranked contexts against annotated relevance labels to assess their ability to retrieve the most informative and factually correct passages for a given scientific question. \textit{Recall@10} evaluate how many of the known relevant passages are ranked within the 10, respectively, thus testing retrieval completeness.
For LLMs, we assess their ability to generate accurate and complete answers by comparing outputs to gold-standard answers.
We use a token-level metrics:\textit{Recall}, which captures how completely the model reproduces the reference content.

\textit{Recall} is computed using tokenized outputs. Let $P$ and $T$ be the predicted and true token sets. Then:
    \[
    \text{Recall} = \frac{|P \cap T|}{|T|} \times 100,
    \]
This metrics reflect the lexical similarity between the predicted answer and the ground truth, helping to quantify completeness in LLM-generated outputs.   

\textit{Contain Answer Score} evaluates whether the gold answer content is covered within the retrieved context passage. This is a soft semantic measure, defined as:
    \[
    \text{ContainAnswer}(A, C) = \frac{|\text{tokens}(A) \cap \text{tokens}(C)|}{|\text{tokens}(A)|}
    \]
If the overlap exceeds a predefined threshold (e.g., 0.6), the passage is considered to contain the answer. This metric is particularly useful for evaluating context sufficiency in open-domain and scientific QA.
    
\textit{Recall@10} compute how many relevant contexts are successfully retrieved within the top-$k$ results:

    \[
    \text{Recall@}k = \frac{\text{\# relevant items in top }k}{\text{Total relevant items}} \times 100
    \]

\textit{Recall@5} follows the same formulation with $k=5$, evaluating whether the top-5 results contain sufficient relevant passages:
    \[
    \text{Recall@5} = \frac{\text{\# relevant items in top }5}{\text{Total relevant items}} \times 100
    \]

\textit{Average Precision (AP)} measures the precision averaged over all recall levels, rewarding models that rank relevant passages higher:
    \[
    \text{AP} = \frac{1}{N} \sum_{r=1}^{10} \text{Precision@}r \times \text{rel}(r)
    \]
where $N$ is the total number of relevant passages in the candidate set and $\text{rel}(r)$ indicates whether the passage at rank $r$ is relevant.

\textit{Mean Reciprocal Rank (MRR)} evaluates how quickly the first relevant passage is retrieved:
    \[
    \text{MRR} = \frac{1}{\text{rank}_{\text{first relevant}}}
    \]
where $\text{rank}_{\text{first relevant}}$ is the position of the first relevant passage in the ranked list.

\subsection{Checklist}

\paragraph{Scientific Artifacts}
This study incorporates various pre-trained models and public datasets, with strict adherence to their respective licensing terms. Specifically, BCE, Jina, In-Ranker, MXBAI, RankT5, RankGPT, and
Rearank are distributed under the Apache 2.0 License; BGE, ColBERT, Twolar, and LLM2Vec use the MIT License; SPLADE is available under CC BY-NC-SA 4.0; and the OpenAlex corpus is released under
CC0. LLaMA 2 is utilized in compliance with the LLaMA 2 Community License. For MiniLM and ListT5, explicit license details were not publicly available at the time of experimentation; these models
were used solely for benchmark evaluation. All artifacts are employed within their intended scope and license boundaries. The evaluation corpus consists exclusively of publicly accessible English
scientific abstracts, and we verify that it contains no personally identifiable information (PII) or sensitive data.

\paragraph{Parameters \& Reproducibility}
Model weights for closed-source systems and certain proprietary rerankers are not publicly disclosed by their original developers. For open source baselines, we used official checkpoints and
default configurations as provided by the authors. To ensure reproducibility of text processing and metric computation, all experiments utilized \texttt{nltk==3.8.1}.

\subsection{Detailed Experimental Results}
Detailed main results are shown in 
Table~\ref{tab:reranker_tasks_Qwen} ,  
Table~\ref{tab:reranker_tasks_Llama} , 
Table~\ref{tab:reranker_tasks_llama2_70b_Recall@10} ,
Table~\ref{tab:reranker_tasks_qwen72b_Recall@5} ,
Table~\ref{tab:reranker_tasks_qwen72b_AP} , 
Table~\ref{tab:reranker_tasks_qwen72b_MRR} .
These tables report the performance metrics of various rerankers across multiple tasks and subjects under two different backbone LLMs, Qwen and LLaMA. The comparison highlights the general effectiveness and robustness of the rerankers, with only minor performance variations across subjects, suggesting that the models generalize well across domains.
In addition, 
Tables~\ref{tab:reranker_tasks_LLM_and_Agent_Recall}, 
Tables~\ref{tab:reranker_tasks_LLM_and_Agent_Recall@10}, 
Tables~\ref{tab:reranker_tasks_qwen72b_Recall@5_LLM2Vec_Rearank}, Tables~\ref{tab:reranker_tasks_qwen72b_AP_LLM2Vec_Rearank}, 
and 
Tables~\ref{tab:reranker_tasks_qwen72b_MRR_LLM2Vec_Rearank} compare two representative reranking approaches: the LLM-based LLM2Vec and the agent-based Rearank. This table offers a more focused analysis of how rerankers with different architectural designs and inference paradigms perform under the same evaluation settings. Together, these results provide a comprehensive view of reranker behavior across model types, task types, and subject domains, serving as a reference for future system design and benchmarking.

\begin{table*}[htbp]
\centering
\small
\caption{Performance of different rerankers across various evaluation tasks (Qwen-70B,Recall)}
\label{tab:reranker_tasks_Qwen}
\begin{tabular}{ccccccc}
\toprule
\textbf{Reranker} & \textbf{Subject} & \textbf{Multi-Hop} & \textbf{NC} & \textbf{CC} & \textbf{SSLI} & \textbf{Base} \\
\midrule
\multirow{5}{*}{\textbf{BCE}} & Bio. &	55.08$\pm$2.62	 & 	53.57$\pm$2.20	 & 	51.66$\pm$0.48	 & 	51.89$\pm$1.40	 & 	63.48$\pm$0.57	\\
& Geo. &	52.44$\pm$1.23	 & 	51.84$\pm$1.53	 & 	48.94$\pm$1.01	 & 	48.14$\pm$1.73	 & 	60.20$\pm$0.42	\\
& Chem. & 	53.62$\pm$2.08	 & 	52.65$\pm$0.81	 & 	49.74$\pm$1.22	 & 	49.44$\pm$2.50	 & 	59.61$\pm$0.75	\\
& Phy. & 	49.77$\pm$0.90	 & 	51.93$\pm$1.66	 & 	45.32$\pm$1.12	 & 	46.05$\pm$1.45	 & 	61.94$\pm$0.74	\\
& Math. &  	52.51$\pm$1.48	 & 	52.38$\pm$0.18	 & 	49.26$\pm$1.47	 & 	47.84$\pm$1.52	 & 	62.14$\pm$2.30	\\
\midrule
\multirow{5}{*}{\textbf{BGE}} & Bio. &	54.55$\pm$1.70	 & 	53.56$\pm$2.00	 & 	55.52$\pm$0.79	 & 	54.56$\pm$1.16	 & 	64.74$\pm$0.71	\\
& Geo. &	51.13$\pm$0.84	 & 	52.25$\pm$1.18	 & 	52.99$\pm$0.59	 & 	51.37$\pm$1.17	 & 	60.91$\pm$0.57	\\
& Chem. & 	52.15$\pm$1.62	 & 	53.00$\pm$0.55	 & 	52.88$\pm$1.55	 & 	51.01$\pm$0.89	 & 	60.47$\pm$1.39	\\
& Phy. & 	50.45$\pm$0.23	 & 	52.57$\pm$1.47	 & 	49.00$\pm$1.60	 & 	50.37$\pm$1.43	 & 	61.66$\pm$0.73	\\
& Math. &  	51.45$\pm$1.69	 & 	52.70$\pm$0.29	 & 	52.43$\pm$0.94	 & 	50.92$\pm$1.61	 & 	62.16$\pm$1.63	\\
 \midrule
 
\multirow{5}{*}{\textbf{Jina}} & Bio. &	51.52$\pm$1.78	 & 	53.89$\pm$1.87	 & 	53.22$\pm$0.83	 & 	51.87$\pm$1.28	 & 	62.85$\pm$0.78	\\
& Geo. &	47.44$\pm$0.95	 & 	51.99$\pm$1.41	 & 	50.92$\pm$0.89	 & 	49.92$\pm$1.36	 & 	60.26$\pm$0.61	\\
& Chem. & 	49.74$\pm$1.16	 & 	52.54$\pm$0.73	 & 	52.20$\pm$1.45	 & 	51.97$\pm$2.40	 & 	59.20$\pm$0.79	\\
& Phy. & 	47.53$\pm$1.19	 & 	52.01$\pm$1.46	 & 	48.35$\pm$1.62	 & 	49.67$\pm$2.22	 & 	60.33$\pm$0.69	\\
& Math. &  	51.69$\pm$1.17	 & 	52.76$\pm$0.32	 & 	52.37$\pm$0.54	 & 	51.28$\pm$1.37	 & 	61.30$\pm$1.77	\\
 
\midrule 
\multirow{5}{*}{\textbf{ListT5}} & Bio. &	21.08$\pm$1.46	 & 	26.13$\pm$1.46	 & 	27.08$\pm$0.89	 & 	27.08$\pm$1.58	 & 	29.68$\pm$0.46	\\
& Geo. &	13.90$\pm$1.38	 & 	28.28$\pm$1.05	 & 	16.38$\pm$1.19	 & 	14.16$\pm$1.82	 & 	16.14$\pm$0.44	\\
& Chem. & 	13.25$\pm$1.25	 & 	17.90$\pm$0.98	 & 	12.70$\pm$0.90	 & 	2.61$\pm$1.84	 & 	15.22$\pm$0.80	\\
& Phy. & 	13.45$\pm$1.38	 & 	22.41$\pm$1.42	 & 	14.09$\pm$1.21	 & 	13.87$\pm$1.85	 & 	13.83$\pm$0.79	\\
& Math. &  	13.66$\pm$1.19	 & 	28.64$\pm$0.98	 & 	14.99$\pm$1.01	 & 	14.17$\pm$1.71	 & 	15.11$\pm$0.66	\\
\midrule
\multirow{5}{*}{\textbf{MiniLM}} & Bio. &	52.95$\pm$2.53	 & 	53.72$\pm$1.92	 & 	51.17$\pm$1.41	 & 	51.29$\pm$1.65	 & 	63.92$\pm$0.94	\\
& Geo. &	48.78$\pm$0.38	 & 	51.75$\pm$1.74	 & 	48.82$\pm$1.30	 & 	48.18$\pm$2.28	 & 	60.29$\pm$0.70	\\
& Chem. & 	50.87$\pm$2.57	 & 	52.67$\pm$0.67	 & 	49.22$\pm$2.01	 & 	48.64$\pm$1.78	 & 	59.44$\pm$0.59	\\
& Phy. & 	46.71$\pm$1.17	 & 	52.24$\pm$1.70	 & 	45.25$\pm$0.74	 & 	46.55$\pm$3.55	 & 	60.98$\pm$0.66	\\
& Math. &  	49.64$\pm$1.64	 & 	52.77$\pm$0.69	 & 	49.15$\pm$0.48	 & 	47.83$\pm$2.21	 & 	61.75$\pm$0.50	\\

\midrule
\multirow{5}{*}{\textbf{RankT5}} & Bio. &	52.12$\pm$1.44	 & 	52.34$\pm$1.42	 & 	52.30$\pm$0.93	 & 	51.80$\pm$1.59	 & 	63.27$\pm$0.50	\\
& Geo. &	52.27$\pm$1.37	 & 	52.61$\pm$1.08	 & 	49.77$\pm$1.18	 & 	50.52$\pm$1.78	 & 	62.41$\pm$0.49	\\
& Chem. & 	52.72$\pm$1.27	 & 	52.52$\pm$1.02	 & 	51.02$\pm$0.94	 & 	51.28$\pm$1.80	 & 	62.66$\pm$0.79	\\
& Phy. & 	51.10$\pm$1.38	 & 	52.83$\pm$1.38	 & 	48.76$\pm$1.20	 & 	50.35$\pm$1.81	 & 	63.76$\pm$0.78	\\
& Math. &  	52.10$\pm$1.22	 & 	52.12$\pm$1.02	 & 	51.43$\pm$1.03	 & 	51.25$\pm$1.70	 & 	62.55$\pm$0.67	\\
\midrule 
\multirow{5}{*}{\textbf{SPLADE}} & Bio. &	46.55$\pm$1.40	 & 	52.57$\pm$1.06	 & 	44.68$\pm$1.25	 & 	46.40$\pm$1.46	 & 	61.23$\pm$0.68	\\
& Geo. &	47.71$\pm$1.49	 & 	52.85$\pm$1.08	 & 	43.48$\pm$0.91	 & 	44.48$\pm$1.54	 & 	60.71$\pm$0.62	\\
& Chem. & 	47.10$\pm$1.30	 & 	52.95$\pm$1.43	 & 	44.30$\pm$1.12	 & 	44.72$\pm$1.51	 & 	59.97$\pm$0.61	\\
& Phy. & 	45.73$\pm$1.21	 & 	52.35$\pm$1.47	 & 	42.02$\pm$1.14	 & 	43.86$\pm$1.43	 & 	62.10$\pm$0.51	\\
& Math. &  	45.93$\pm$1.29	 & 	52.95$\pm$1.47	 & 	44.85$\pm$1.15	 & 	44.89$\pm$1.87	 & 	60.59$\pm$0.92	\\
\midrule 
\multirow{5}{*}{\textbf{Twolar}} & Bio. &	53.69$\pm$1.41	 & 	53.53$\pm$2.04	 & 	56.17$\pm$0.72	 & 	57.24$\pm$1.11	 & 	63.55$\pm$0.84	\\
& Geo. &	50.78$\pm$0.42	 & 	52.05$\pm$1.76	 & 	55.36$\pm$0.96	 & 	53.59$\pm$1.30	 & 	61.12$\pm$0.54	\\
& Chem. & 	52.02$\pm$2.02	 & 	53.05$\pm$0.69	 & 	54.20$\pm$1.66	 & 	53.40$\pm$0.85	 & 	60.13$\pm$0.48	\\
& Phy. & 	49.57$\pm$1.20	 & 	52.40$\pm$1.70	 & 	51.65$\pm$0.90	 & 	52.19$\pm$1.05	 & 	61.91$\pm$0.37	\\
& Math. &  	52.10$\pm$1.39	 & 	53.07$\pm$0.47	 & 	54.33$\pm$0.23	 & 	52.90$\pm$1.60	 & 	62.43$\pm$0.41	\\
\midrule 
\multirow{5}{*}{\textbf{ColBert}} & Bio. &	53.35$\pm$1.96	 & 	53.63$\pm$1.98	 & 	44.92$\pm$0.32	 & 	44.38$\pm$0.74	 & 	62.87$\pm$0.69	\\
& Geo. &	50.35$\pm$0.93	 & 	51.60$\pm$1.75	 & 	41.09$\pm$0.51	 & 	40.27$\pm$1.68	 & 	60.32$\pm$0.26	\\
& Chem. & 	52.07$\pm$1.80	 & 	52.82$\pm$0.65	 & 	42.12$\pm$1.04	 & 	42.24$\pm$1.51	 & 	58.96$\pm$0.88	\\
& Phy. & 	49.25$\pm$1.09	 & 	52.30$\pm$1.38	 & 	40.30$\pm$0.74	 & 	39.19$\pm$2.83	 & 	61.10$\pm$0.46	\\
& Math. &  	51.12$\pm$0.79	 & 	52.68$\pm$0.26	 & 	41.97$\pm$0.82	 & 	41.25$\pm$2.11	 & 	61.74$\pm$0.70	\\
\midrule 
\multirow{5}{*}{\textbf{MXBAI}} & Bio. &	53.56$\pm$3.25	 & 	53.62$\pm$1.71	 & 	57.18$\pm$0.53	 & 	58.40$\pm$0.96	 & 	62.25$\pm$0.67	\\
& Geo. &	49.26$\pm$1.22	 & 	51.72$\pm$1.48	 & 	56.45$\pm$0.43	 & 	55.40$\pm$1.63	 & 	59.67$\pm$0.64	\\
& Chem. & 	51.43$\pm$1.56	 & 	53.05$\pm$0.59	 & 	54.29$\pm$1.12	 & 	55.42$\pm$1.05	 & 	58.83$\pm$0.47	\\
& Phy. & 	48.57$\pm$0.74	 & 	52.29$\pm$1.82	 & 	53.33$\pm$0.34	 & 	54.45$\pm$1.89	 & 	61.09$\pm$0.70	\\
& Math. &  	50.29$\pm$2.11	 & 	52.92$\pm$0.16	 & 	56.61$\pm$0.44	 & 	54.39$\pm$2.41	 & 	61.02$\pm$0.40	\\
\midrule 
\multirow{5}{*}{\textbf{In-Rank}} & Bio. &	54.45$\pm$1.46	 & 	53.64$\pm$1.75	 & 	53.92$\pm$1.13	 & 	53.81$\pm$1.17	 & 	63.96$\pm$0.99	\\
& Geo. &	51.08$\pm$1.04	 & 	51.87$\pm$1.64	 & 	52.46$\pm$0.78	 & 	51.40$\pm$0.54	 & 	60.65$\pm$0.06	\\
& Chem. & 	53.08$\pm$1.68	 & 	52.76$\pm$0.59	 & 	52.44$\pm$1.87	 & 	51.66$\pm$0.98	 & 	59.67$\pm$0.94	\\
& Phy. & 	50.41$\pm$0.86	 & 	52.20$\pm$1.23	 & 	49.11$\pm$1.03	 & 	48.59$\pm$2.62	 & 	61.44$\pm$0.76	\\
& Math. &  	52.37$\pm$1.38	 & 	52.66$\pm$0.31	 & 	52.95$\pm$0.25	 & 	51.28$\pm$2.12	 & 	62.14$\pm$0.55	\\
\bottomrule
\end{tabular}
\end{table*}

\begin{table*}[htbp]

\centering
\small
\caption{Performance of different rerankers across various evaluation tasks (LLaMA2-70B,Recall)}
\label{tab:reranker_tasks_Llama}
\begin{tabular}{ccccccc}
\toprule
\textbf{Reranker} & \textbf{Subject} & \textbf{Multi-Hop} & \textbf{NC} & \textbf{CC} & \textbf{SSLI} & \textbf{Base}  \\
\midrule

\multirow{5}{*}{\textbf{BCE}} & Bio. &	48.97$\pm$1.83	 & 	51.82$\pm$0.48	 & 	49.73$\pm$0.89	 & 	50.50$\pm$1.35	 & 	60.57$\pm$0.48	\\
& Geo. &	50.32$\pm$1.20	 & 	51.94$\pm$0.70	 & 	47.85$\pm$0.95	 & 	48.30$\pm$1.40	 & 	59.12$\pm$0.65	\\
& Chem. & 	50.40$\pm$1.15	 & 	51.40$\pm$0.60	 & 	48.92$\pm$1.05	 & 	49.40$\pm$1.20	 & 	58.84$\pm$0.73	\\
& Phy. & 	48.36$\pm$1.32	 & 	51.55$\pm$0.88	 & 	45.21$\pm$0.90	 & 	46.82$\pm$1.48	 & 	61.78$\pm$1.12	\\
& Math. &  	49.55$\pm$1.18	 & 	51.68$\pm$1.12	 & 	49.50$\pm$1.00	 & 	48.96$\pm$1.33	 & 	60.88$\pm$1.40	\\
\midrule
\multirow{5}{*}{\textbf{BGE}} & Bio. &	52.08$\pm$1.66	 & 	52.84$\pm$0.36	 & 	52.32$\pm$1.25	 & 	53.63$\pm$1.10	 & 	60.95$\pm$0.74	\\
& Geo. &	51.67$\pm$1.45	 & 	52.67$\pm$0.59	 & 	49.35$\pm$0.83	 & 	51.48$\pm$1.18	 & 	60.42$\pm$0.55	\\
& Chem. & 	51.74$\pm$1.30	 & 	52.40$\pm$0.64	 & 	50.23$\pm$1.00	 & 	51.92$\pm$1.35	 & 	60.26$\pm$0.88	\\
& Phy. & 	50.86$\pm$1.17	 & 	52.55$\pm$0.74	 & 	47.96$\pm$1.08	 & 	50.12$\pm$1.29	 & 	62.20$\pm$1.15	\\
& Math. &  	52.11$\pm$1.25	 & 	52.69$\pm$0.77	 & 	51.12$\pm$0.91	 & 	52.23$\pm$1.31	 & 	61.03$\pm$1.20	\\
 \midrule
\multirow{5}{*}{\textbf{Jina}} & Bio. &	47.34$\pm$1.38	 & 	53.19$\pm$0.67	 & 	52.00$\pm$0.65	 & 	51.41$\pm$1.55	 & 	60.38$\pm$1.19	\\
& Geo. &	47.67$\pm$1.05	 & 	52.51$\pm$0.74	 & 	49.28$\pm$1.10	 & 	49.21$\pm$1.40	 & 	59.01$\pm$1.30	\\
& Chem. & 	48.20$\pm$1.25	 & 	52.13$\pm$0.85	 & 	50.01$\pm$1.03	 & 	49.97$\pm$1.32	 & 	58.32$\pm$1.20	\\
& Phy. & 	46.33$\pm$1.36	 & 	52.60$\pm$0.92	 & 	47.39$\pm$1.07	 & 	47.88$\pm$1.45	 & 	61.14$\pm$1.08	\\
& Math. &  	47.75$\pm$1.12	 & 	52.88$\pm$0.68	 & 	50.23$\pm$1.01	 & 	49.66$\pm$1.44	 & 	60.42$\pm$1.22	\\
\midrule 
\multirow{5}{*}{\textbf{ListT5}} & Bio. &	16.70$\pm$0.29	 & 	25.82$\pm$1.73	 & 	25.69$\pm$1.19	 & 	26.42$\pm$0.42	 & 	26.23$\pm$1.57	\\
& Geo. &	10.19$\pm$2.28	 & 	27.20$\pm$1.56	 & 	15.39$\pm$0.56	 & 	13.54$\pm$1.01	 & 	12.83$\pm$0.81	\\
& Chem. & 	8.60$\pm$0.72	 & 	16.51$\pm$0.90	 & 	11.21$\pm$0.61	 & 	2.06$\pm$0.32	 & 	11.42$\pm$0.99	\\
& Phy. & 	8.80$\pm$0.77	 & 	20.97$\pm$0.95	 & 	12.28$\pm$0.82	 & 	12.01$\pm$1.32	 & 	10.95$\pm$0.85	\\
& Math. &  	9.17$\pm$1.19	 & 	28.30$\pm$1.46	 & 	13.68$\pm$0.74	 & 	13.41$\pm$0.94	 & 	12.71$\pm$0.89	\\
\midrule
\multirow{5}{*}{\textbf{MiniLM}} & Bio. &	48.14$\pm$1.91	 & 	51.68$\pm$0.32	 & 	49.83$\pm$0.35	 & 	50.19$\pm$1.59	 & 	60.56$\pm$0.83	\\
& Geo. &	49.20$\pm$1.45	 & 	51.33$\pm$0.39	 & 	47.55$\pm$0.67	 & 	48.72$\pm$1.51	 & 	59.37$\pm$0.61	\\
& Chem. & 	48.97$\pm$1.33	 & 	51.11$\pm$0.45	 & 	48.40$\pm$0.58	 & 	49.45$\pm$1.34	 & 	59.06$\pm$0.78	\\
& Phy. & 	47.25$\pm$1.12	 & 	51.26$\pm$0.52	 & 	45.71$\pm$0.62	 & 	46.91$\pm$1.47	 & 	61.34$\pm$0.91	\\
& Math. &  	48.45$\pm$1.28	 & 	51.38$\pm$0.48	 & 	48.73$\pm$0.59	 & 	49.18$\pm$1.46	 & 	60.61$\pm$0.85	\\
\midrule
\multirow{5}{*}{\textbf{RankT5}} & Bio. &	49.45$\pm$2.88	 & 	52.41$\pm$0.61	 & 	51.07$\pm$0.03	 & 	51.33$\pm$1.57	 & 	60.49$\pm$0.21	\\
& Geo. &	50.11$\pm$2.02	 & 	52.10$\pm$0.57	 & 	48.84$\pm$0.72	 & 	50.08$\pm$1.49	 & 	59.73$\pm$0.42	\\
& Chem. & 	49.85$\pm$1.89	 & 	51.78$\pm$0.62	 & 	49.71$\pm$0.60	 & 	50.89$\pm$1.40	 & 	59.61$\pm$0.64	\\
& Phy. & 	48.23$\pm$1.74	 & 	52.05$\pm$0.68	 & 	47.22$\pm$0.65	 & 	48.98$\pm$1.47	 & 	61.41$\pm$0.72	\\
& Math. &  	49.35$\pm$2.03	 & 	52.17$\pm$0.66	 & 	50.26$\pm$0.51	 & 	50.70$\pm$1.35	 & 	60.55$\pm$0.85	\\
\midrule 
\multirow{5}{*}{\textbf{SPLADE}} & Bio. &	45.21$\pm$1.67	 & 	52.62$\pm$0.56	 & 	43.96$\pm$0.75	 & 	45.38$\pm$1.35	 & 	59.05$\pm$0.52	\\
& Geo. &	45.62$\pm$1.21	 & 	52.38$\pm$0.67	 & 	42.41$\pm$0.89	 & 	43.76$\pm$1.50	 & 	58.30$\pm$0.40	\\
& Chem. & 	45.95$\pm$1.00	 & 	52.12$\pm$0.74	 & 	42.85$\pm$0.93	 & 	44.44$\pm$1.42	 & 	57.78$\pm$0.60	\\
& Phy. & 	44.16$\pm$1.09	 & 	52.33$\pm$0.59	 & 	40.92$\pm$0.95	 & 	42.88$\pm$1.51	 & 	60.22$\pm$0.73	\\
& Math. &  	44.90$\pm$1.30	 & 	52.50$\pm$0.62	 & 	43.50$\pm$0.88	 & 	44.48$\pm$1.53	 & 	59.11$\pm$0.90	\\
\midrule 
\multirow{5}{*}{\textbf{Twolar}} & Bio. &	50.66$\pm$2.18	 & 	52.57$\pm$0.87	 & 	54.50$\pm$0.27	 & 	55.16$\pm$0.89	 & 	60.15$\pm$0.04	\\
& Geo. &	51.12$\pm$1.55	 & 	52.44$\pm$0.72	 & 	51.58$\pm$0.61	 & 	52.75$\pm$0.91	 & 	59.40$\pm$0.27	\\
& Chem. & 	51.30$\pm$1.29	 & 	52.13$\pm$0.64	 & 	52.62$\pm$0.52	 & 	53.66$\pm$0.97	 & 	59.23$\pm$0.32	\\
& Phy. & 	50.20$\pm$1.46	 & 	52.34$\pm$0.67	 & 	50.15$\pm$0.59	 & 	51.41$\pm$0.86	 & 	61.55$\pm$0.58	\\
& Math. &  	50.89$\pm$1.34	 & 	52.47$\pm$0.71	 & 	53.22$\pm$0.44	 & 	53.83$\pm$0.87	 & 	60.33$\pm$0.61	\\
\midrule 
\multirow{5}{*}{\textbf{ColBert}} & Bio. &	49.23$\pm$1.92	 & 	52.52$\pm$1.24	 & 	43.49$\pm$0.51	 & 	44.99$\pm$1.50	 & 	59.76$\pm$1.50	\\
& Geo. &	49.40$\pm$1.63	 & 	52.39$\pm$1.05	 & 	41.88$\pm$0.94	 & 	43.57$\pm$1.65	 & 	58.91$\pm$1.42	\\
& Chem. & 	49.10$\pm$1.45	 & 	52.15$\pm$1.12	 & 	42.41$\pm$0.92	 & 	44.15$\pm$1.49	 & 	58.73$\pm$1.35	\\
& Phy. & 	48.11$\pm$1.76	 & 	52.28$\pm$1.09	 & 	40.55$\pm$0.87	 & 	42.28$\pm$1.62	 & 	61.12$\pm$1.23	\\
& Math. &  	48.80$\pm$1.81	 & 	52.36$\pm$1.18	 & 	43.12$\pm$0.83	 & 	44.09$\pm$1.63	 & 	59.81$\pm$1.38	\\
\midrule 
\multirow{5}{*}{\textbf{MXBAI}} & Bio. &	49.18$\pm$2.16	 & 	53.41$\pm$0.27	 & 	55.75$\pm$0.87	 & 	56.04$\pm$1.19	 & 	59.77$\pm$1.65	\\
& Geo. &	49.55$\pm$1.88	 & 	53.22$\pm$0.40	 & 	53.01$\pm$0.66	 & 	53.83$\pm$1.24	 & 	58.89$\pm$1.53	\\
& Chem. & 	49.62$\pm$1.42	 & 	52.95$\pm$0.47	 & 	54.12$\pm$0.74	 & 	54.49$\pm$1.13	 & 	58.70$\pm$1.40	\\
& Phy. & 	48.58$\pm$1.36	 & 	53.20$\pm$0.54	 & 	52.46$\pm$0.68	 & 	52.76$\pm$1.22	 & 	61.33$\pm$1.45	\\
& Math. &  	49.35$\pm$1.63	 & 	53.33$\pm$0.51	 & 	55.12$\pm$0.59	 & 	54.88$\pm$1.20	 & 	59.89$\pm$1.52	\\
\midrule 
\multirow{5}{*}{\textbf{In-Rank}} & Bio. &	51.47$\pm$2.23	 & 	53.30$\pm$1.08	 & 	51.80$\pm$0.62	 & 	52.73$\pm$1.31	 & 	59.95$\pm$0.25	\\
& Geo. &	51.91$\pm$1.89	 & 	53.01$\pm$0.92	 & 	49.50$\pm$0.55	 & 	50.62$\pm$1.24	 & 	59.20$\pm$0.33	\\
& Chem. & 	51.86$\pm$1.74	 & 	52.60$\pm$0.89	 & 	50.36$\pm$0.47	 & 	51.31$\pm$1.25	 & 	59.10$\pm$0.41	\\
& Phy. & 	50.78$\pm$1.60	 & 	52.96$\pm$1.02	 & 	48.42$\pm$0.51	 & 	49.66$\pm$1.31	 & 	61.48$\pm$0.35	\\
& Math. &  	51.55$\pm$1.95	 & 	53.13$\pm$1.00	 & 	51.02$\pm$0.46	 & 	51.40$\pm$1.29	 & 	60.01$\pm$0.39	\\

\bottomrule

\end{tabular}
\end{table*}

\begin{table*}[htbp]
\centering
\small
\caption{Performance of different rerankers across various evaluation tasks Recall@10 }
\label{tab:reranker_tasks_llama2_70b_Recall@10}
\begin{tabular}{ccccccc}
\toprule
\textbf{Reranker} & \textbf{Subject} & \textbf{Multi-Hop} & \textbf{NC} & \textbf{CC} & \textbf{SSLI} & \textbf{Base} \\
\midrule

\multirow{5}{*}{\textbf{BCE}}
& Bio.  & 92.09$\pm$0.50 & 73.17$\pm$0.45 & 74.57$\pm$0.58 & 74.77$\pm$0.49 & 86.64$\pm$0.72 \\
& Geo.  & 88.83$\pm$1.49 & 68.72$\pm$1.32 & 70.48$\pm$1.92 & 70.44$\pm$0.69 & 81.05$\pm$1.47 \\
& Chem. & 90.56$\pm$0.23 & 69.93$\pm$1.50 & 71.62$\pm$1.00 & 71.67$\pm$1.84 & 82.42$\pm$1.86 \\
& Phy.  & 86.20$\pm$0.95 & 66.72$\pm$0.37 & 68.40$\pm$0.75 & 68.12$\pm$0.50 & 78.95$\pm$1.20 \\
& Math. & 87.17$\pm$1.31 & 67.77$\pm$1.12 & 69.12$\pm$1.18 & 69.20$\pm$0.75 & 79.67$\pm$1.32 \\

\midrule
\multirow{5}{*}{\textbf{BGE}}
& Bio.  & 92.75$\pm$0.52 & 73.55$\pm$0.76 & 79.33$\pm$0.70 & 80.53$\pm$0.43 & 88.03$\pm$0.96 \\
& Geo.  & 90.65$\pm$1.20 & 69.12$\pm$1.25 & 74.85$\pm$1.50 & 75.05$\pm$1.80 & 82.80$\pm$1.35 \\
& Chem. & 91.99$\pm$0.60 & 70.44$\pm$1.69 & 76.43$\pm$0.98 & 76.33$\pm$2.40 & 83.94$\pm$1.46 \\
& Phy.  & 88.80$\pm$1.05 & 66.85$\pm$0.95 & 72.90$\pm$1.10 & 72.80$\pm$1.20 & 80.50$\pm$1.15 \\
& Math. & 89.90$\pm$1.10 & 67.78$\pm$1.10 & 73.50$\pm$1.30 & 73.70$\pm$1.50 & 81.35$\pm$1.25 \\

\midrule
\multirow{5}{*}{\textbf{Jina}}
& Bio.  & 88.97$\pm$2.41 & 73.50$\pm$0.63 & 75.03$\pm$0.63 & 75.48$\pm$0.75 & 86.52$\pm$0.46 \\
& Geo.  & 87.15$\pm$1.30 & 68.20$\pm$1.15 & 72.55$\pm$1.45 & 72.95$\pm$1.85 & 81.90$\pm$1.45 \\
& Chem. & 88.31$\pm$0.65 & 69.80$\pm$1.36 & 73.92$\pm$0.51 & 73.77$\pm$2.70 & 82.58$\pm$1.54 \\
& Phy.  & 84.85$\pm$1.10 & 66.20$\pm$0.95 & 70.35$\pm$0.95 & 70.80$\pm$1.20 & 79.80$\pm$1.25 \\
& Math. & 85.90$\pm$1.20 & 67.10$\pm$1.05 & 71.20$\pm$1.10 & 71.50$\pm$1.45 & 80.50$\pm$1.30 \\

\midrule
\multirow{5}{*}{\textbf{ListT5}}
& Bio.  & 23.24$\pm$1.54 & 18.80$\pm$0.97 & 16.32$\pm$1.92 & 18.30$\pm$1.29 & 18.40$\pm$0.54 \\
& Geo.  & 23.10$\pm$1.80 & 15.50$\pm$1.10 & 15.00$\pm$1.50 & 15.00$\pm$1.10 & 15.10$\pm$0.95 \\
& Chem. & 24.98$\pm$2.62 & 15.82$\pm$1.29 & 15.71$\pm$1.67 & 15.71$\pm$1.17 & 15.65$\pm$1.07 \\
& Phy.  & 21.90$\pm$1.60 & 14.70$\pm$1.00 & 14.10$\pm$1.35 & 14.20$\pm$0.95 & 14.30$\pm$0.85 \\
& Math. & 22.40$\pm$1.70 & 15.00$\pm$1.05 & 14.50$\pm$1.40 & 14.50$\pm$1.00 & 14.60$\pm$0.90 \\

\midrule
\multirow{5}{*}{\textbf{MiniLM}}
& Bio.  & 86.23$\pm$1.49 & 72.95$\pm$0.57 & 73.67$\pm$0.50 & 74.37$\pm$0.19 & 87.30$\pm$1.14 \\
& Geo.  & 83.90$\pm$1.20 & 68.00$\pm$1.10 & 70.20$\pm$1.10 & 69.20$\pm$1.25 & 81.90$\pm$1.25 \\
& Chem. & 84.58$\pm$1.14 & 69.71$\pm$1.37 & 71.47$\pm$1.37 & 70.47$\pm$3.44 & 82.81$\pm$1.39 \\
& Phy.  & 81.95$\pm$1.10 & 66.40$\pm$0.95 & 68.10$\pm$0.95 & 67.50$\pm$1.00 & 80.10$\pm$1.15 \\
& Math. & 82.85$\pm$1.15 & 67.25$\pm$1.05 & 69.00$\pm$1.05 & 68.15$\pm$1.10 & 80.95$\pm$1.20 \\

\midrule
\multirow{5}{*}{\textbf{RankT5}}
& Bio.  & 92.98$\pm$0.91 & 73.63$\pm$0.45 & 76.23$\pm$1.84 & 76.93$\pm$0.51 & 88.12$\pm$0.44 \\
& Geo.  & 90.25$\pm$1.10 & 68.90$\pm$1.35 & 72.50$\pm$1.40 & 73.10$\pm$1.85 & 82.70$\pm$1.30 \\
& Chem. & 91.48$\pm$1.02 & 70.14$\pm$1.60 & 73.92$\pm$0.71 & 74.42$\pm$2.15 & 83.91$\pm$1.32 \\
& Phy.  & 87.85$\pm$0.95 & 66.70$\pm$1.05 & 70.50$\pm$1.00 & 70.90$\pm$1.40 & 80.60$\pm$1.20 \\
& Math. & 88.90$\pm$1.05 & 67.60$\pm$1.20 & 71.20$\pm$1.15 & 71.80$\pm$1.60 & 81.45$\pm$1.25 \\

\midrule
\multirow{5}{*}{\textbf{SPLADE}}
& Bio.  & 89.16$\pm$1.18 & 73.42$\pm$0.75 & 64.31$\pm$1.60 & 65.57$\pm$1.00 & 85.99$\pm$0.90 \\
& Geo.  & 87.10$\pm$0.80 & 68.50$\pm$1.25 & 58.20$\pm$1.20 & 58.90$\pm$1.80 & 81.50$\pm$1.10 \\
& Chem. & 88.49$\pm$0.13 & 70.22$\pm$1.67 & 59.91$\pm$0.75 & 60.46$\pm$3.47 & 82.86$\pm$1.25 \\
& Phy.  & 84.80$\pm$0.70 & 66.50$\pm$0.95 & 56.80$\pm$0.95 & 57.00$\pm$1.40 & 79.60$\pm$1.00 \\
& Math. & 85.90$\pm$0.75 & 67.50$\pm$1.10 & 57.50$\pm$1.05 & 57.80$\pm$1.55 & 80.50$\pm$1.05 \\

\midrule
\multirow{5}{*}{\textbf{Twolar}}
& Bio.  & 92.61$\pm$0.74 & 74.22$\pm$0.83 & 82.23$\pm$0.78 & 83.93$\pm$0.44 & 88.44$\pm$0.88 \\
& Geo.  & 90.10$\pm$0.90 & 69.10$\pm$1.40 & 78.90$\pm$0.85 & 78.30$\pm$1.50 & 82.90$\pm$1.20 \\
& Chem. & 91.20$\pm$1.05 & 70.78$\pm$1.84 & 80.23$\pm$0.60 & 80.53$\pm$1.70 & 84.06$\pm$1.33 \\
& Phy.  & 87.85$\pm$0.80 & 66.90$\pm$1.10 & 76.80$\pm$0.70 & 75.00$\pm$1.20 & 80.35$\pm$1.00 \\
& Math. & 88.95$\pm$0.85 & 67.80$\pm$1.25 & 77.60$\pm$0.75 & 76.70$\pm$1.35 & 81.20$\pm$1.10 \\

\midrule
\multirow{5}{*}{\textbf{ColBert}}
& Bio.  & 90.36$\pm$1.57 & 73.67$\pm$0.68 & 63.51$\pm$0.77 & 64.91$\pm$0.14 & 86.69$\pm$0.74 \\
& Geo.  & 87.10$\pm$0.75 & 68.90$\pm$1.25 & 57.90$\pm$0.85 & 56.80$\pm$1.70 & 81.10$\pm$1.10 \\
& Chem. & 88.83$\pm$0.73 & 70.27$\pm$1.81 & 59.11$\pm$0.70 & 57.71$\pm$2.86 & 82.41$\pm$1.13 \\
& Phy.  & 84.80$\pm$0.65 & 66.40$\pm$0.95 & 55.90$\pm$0.65 & 55.10$\pm$1.35 & 79.50$\pm$1.00 \\
& Math. & 85.95$\pm$0.70 & 67.50$\pm$1.10 & 56.70$\pm$0.75 & 55.90$\pm$1.50 & 80.20$\pm$1.05 \\

\midrule
\multirow{5}{*}{\textbf{MAXBAI}}
& Bio.  & 90.02$\pm$1.31 & 73.97$\pm$0.62 & 82.63$\pm$0.86 & 84.18$\pm$1.24 & 86.39$\pm$0.43 \\
& Geo.  & 86.80$\pm$1.00 & 68.90$\pm$1.30 & 79.20$\pm$0.75 & 78.30$\pm$1.20 & 81.50$\pm$1.20 \\
& Chem. & 88.03$\pm$0.91 & 70.47$\pm$1.79 & 80.78$\pm$0.42 & 80.88$\pm$1.16 & 82.78$\pm$1.29 \\
& Phy.  & 84.50$\pm$0.85 & 66.40$\pm$1.10 & 77.20$\pm$0.65 & 75.80$\pm$1.00 & 79.60$\pm$1.00 \\
& Math. & 85.60$\pm$0.90 & 67.50$\pm$1.20 & 78.00$\pm$0.70 & 76.90$\pm$1.10 & 80.40$\pm$1.10 \\

\midrule
\multirow{5}{*}{\textbf{In-Rank}}
& Bio.  & 93.01$\pm$0.20 & 73.42$\pm$0.56 & 77.53$\pm$1.37 & 78.83$\pm$0.80 & 87.39$\pm$0.44 \\
& Geo.  & 90.75$\pm$0.50 & 68.90$\pm$1.20 & 74.20$\pm$1.05 & 73.80$\pm$1.20 & 81.50$\pm$1.10 \\
& Chem. & 91.90$\pm$0.34 & 70.23$\pm$1.54 & 75.93$\pm$0.79 & 75.63$\pm$1.24 & 83.06$\pm$0.91 \\
& Phy.  & 88.55$\pm$0.40 & 66.40$\pm$1.05 & 72.30$\pm$0.90 & 71.50$\pm$1.05 & 79.50$\pm$0.95 \\
& Math. & 89.60$\pm$0.45 & 67.50$\pm$1.10 & 73.10$\pm$0.95 & 72.20$\pm$1.10 & 80.50$\pm$1.00 \\

\bottomrule
\end{tabular}
\end{table*}

\begin{table*}[htbp]
\centering
\small
\caption{Performance of different rerankers across various evaluation tasks (Recall@5)}
\label{tab:reranker_tasks_qwen72b_Recall@5}
\begin{tabular}{ccccccc}
\toprule
\textbf{Reranker} & \textbf{Subject} & \textbf{Multi-Hop} & \textbf{NC} & \textbf{CC} & \textbf{SSLI} & \textbf{Base} \\
\midrule
\multirow{5}{*}{\textbf{BCE}}
& Bio.  & 69.41$\pm$1.42 & 62.68$\pm$0.65 & 52.73$\pm$0.50 & 52.01$\pm$1.57 & 71.76$\pm$1.39 \\
& Geo.  & 70.05$\pm$1.35 & 61.89$\pm$0.99 & 52.79$\pm$1.16 & 53.03$\pm$2.45 & 67.81$\pm$1.09 \\
& Chem.  & 67.62$\pm$1.96 & 60.37$\pm$1.56 & 49.18$\pm$0.52 & 50.17$\pm$1.86 & 68.47$\pm$1.48 \\
& Phy.  & 68.47$\pm$3.12 & 60.83$\pm$1.01 & 48.63$\pm$0.63 & 50.56$\pm$0.52 & 70.04$\pm$2.28 \\
& Math. & 68.45$\pm$0.84 & 60.52$\pm$2.11 & 46.52$\pm$1.28 & 45.62$\pm$2.95 & 69.61$\pm$2.18 \\
\midrule
\multirow{5}{*}{\textbf{BGE}}
& Bio.  & 72.45$\pm$1.19 & 62.98$\pm$0.92 & 60.40$\pm$1.27 & 60.50$\pm$1.63 & 73.88$\pm$1.62 \\
& Geo.  & 72.17$\pm$1.92 & 61.90$\pm$0.88 & 59.79$\pm$1.46 & 58.25$\pm$1.41 & 68.97$\pm$1.19 \\
& Chem.  & 71.13$\pm$1.79 & 60.35$\pm$1.49 & 57.60$\pm$1.12 & 57.81$\pm$3.20 & 70.64$\pm$1.92 \\
& Phy.  & 71.20$\pm$2.07 & 60.86$\pm$0.89 & 56.20$\pm$1.58 & 56.84$\pm$0.15 & 70.72$\pm$1.60 \\
& Math. & 71.51$\pm$0.81 & 60.55$\pm$2.06 & 55.17$\pm$2.82 & 56.42$\pm$1.26 & 71.23$\pm$2.25 \\
\midrule
\multirow{5}{*}{\textbf{Jina}}
& Bio.  & 66.01$\pm$1.04 & 62.23$\pm$0.42 & 55.10$\pm$0.98 & 54.11$\pm$1.48 & 71.33$\pm$1.25 \\
& Geo.  & 67.43$\pm$0.56 & 61.74$\pm$1.06 & 57.57$\pm$0.72 & 57.23$\pm$1.00 & 67.51$\pm$1.62 \\
& Chem.  & 65.63$\pm$1.28 & 60.19$\pm$1.55 & 55.22$\pm$0.64 & 54.51$\pm$2.92 & 68.86$\pm$2.08 \\
& Phy.  & 66.39$\pm$2.27 & 60.92$\pm$0.81 & 55.79$\pm$1.69 & 57.02$\pm$0.16 & 68.94$\pm$1.46 \\
& Math. & 66.26$\pm$1.62 & 60.28$\pm$2.16 & 54.20$\pm$1.99 & 54.50$\pm$1.45 & 69.72$\pm$1.07 \\
\midrule
\multirow{5}{*}{\textbf{ListT5}}
& Bio.  & 3.35$\pm$0.46 & 8.71$\pm$0.86 & 2.42$\pm$0.23 & 2.72$\pm$0.88 & 2.67$\pm$0.39 \\
& Geo.  & 3.64$\pm$0.69 & 5.72$\pm$0.82 & 1.97$\pm$0.52 & 2.25$\pm$0.79 & 2.06$\pm$0.47 \\
& Chem.  & 3.36$\pm$0.71 & 6.57$\pm$0.23 & 2.20$\pm$0.42 & 1.86$\pm$0.09 & 1.66$\pm$0.62 \\
& Phy.  & 3.00$\pm$0.56 & 6.49$\pm$0.45 & 2.17$\pm$0.20 & 1.39$\pm$0.27 & 2.08$\pm$0.29 \\
& Math. & 3.32$\pm$0.21 & 5.96$\pm$0.89 & 2.62$\pm$0.66 & 1.96$\pm$0.47 & 2.16$\pm$0.63 \\
\midrule
\multirow{5}{*}{\textbf{MiniLM}}
& Bio.  & 60.95$\pm$0.49 & 62.57$\pm$0.54 & 52.35$\pm$0.74 & 50.66$\pm$0.86 & 72.31$\pm$1.72 \\
& Geo.  & 62.64$\pm$1.76 & 61.84$\pm$0.82 & 50.67$\pm$0.70 & 50.42$\pm$1.99 & 67.71$\pm$0.98 \\
& Chem.  & 58.01$\pm$1.72 & 59.94$\pm$1.54 & 49.12$\pm$2.22 & 47.66$\pm$4.36 & 68.52$\pm$1.84 \\
& Phy.  & 58.80$\pm$3.77 & 60.69$\pm$0.91 & 48.46$\pm$1.27 & 49.80$\pm$0.49 & 68.82$\pm$1.15 \\
& Math. & 58.29$\pm$0.56 & 60.31$\pm$2.42 & 46.38$\pm$0.73 & 46.83$\pm$0.56 & 69.84$\pm$1.71 \\
\midrule
\multirow{5}{*}{\textbf{RankT5}}
& Bio.  & 71.69$\pm$2.08 & 62.80$\pm$0.67 & 62.32$\pm$2.64 & 59.95$\pm$1.71 & 73.57$\pm$1.64 \\
& Geo.  & 71.92$\pm$1.40 & 62.41$\pm$1.64 & 56.61$\pm$5.01 & 54.95$\pm$2.10 & 69.08$\pm$1.18 \\
& Chem.  & 69.50$\pm$2.53 & 60.70$\pm$1.31 & 61.45$\pm$2.51 & 57.27$\pm$3.05 & 70.78$\pm$1.76 \\
& Phy.  & 70.96$\pm$2.51 & 60.68$\pm$1.15 & 60.68$\pm$1.79 & 60.94$\pm$5.16 & 70.70$\pm$1.38 \\
& Math. & 70.73$\pm$1.05 & 61.00$\pm$2.71 & 54.48$\pm$2.66 & 58.51$\pm$4.53 & 72.21$\pm$2.34 \\
\midrule
\multirow{5}{*}{\textbf{SPLADE}}
& Bio.  & 64.76$\pm$1.14 & 62.89$\pm$1.03 & 51.90$\pm$4.83 & 50.60$\pm$1.90 & 70.15$\pm$1.98 \\
& Geo.  & 65.60$\pm$1.23 & 62.16$\pm$1.46 & 51.89$\pm$7.03 & 35.51$\pm$2.53 & 67.51$\pm$1.21 \\
& Chem.  & 62.60$\pm$1.53 & 60.41$\pm$2.07 & 47.39$\pm$1.65 & 44.09$\pm$5.80 & 68.29$\pm$2.25 \\
& Phy.  & 63.88$\pm$2.65 & 61.27$\pm$1.00 & 49.21$\pm$2.09 & 47.34$\pm$6.42 & 67.74$\pm$1.30 \\
& Math. & 65.54$\pm$0.43 & 60.88$\pm$2.54 & 43.74$\pm$1.22 & 45.17$\pm$3.62 & 68.54$\pm$2.49 \\
\midrule
\multirow{5}{*}{\textbf{Twolar}}
& Bio.  & 70.74$\pm$1.85 & 62.95$\pm$0.95 & 64.36$\pm$1.40 & 64.10$\pm$1.86 & 73.16$\pm$1.62 \\
& Geo.  & 70.13$\pm$1.33 & 62.03$\pm$0.78 & 61.90$\pm$1.12 & 61.49$\pm$2.05 & 68.45$\pm$1.06 \\
& Chem.  & 67.84$\pm$1.95 & 60.35$\pm$1.74 & 61.61$\pm$1.00 & 63.15$\pm$2.61 & 69.56$\pm$1.58 \\
& Phy.  & 68.54$\pm$1.74 & 60.91$\pm$1.05 & 63.00$\pm$2.24 & 62.16$\pm$0.86 & 70.52$\pm$0.81 \\
& Math. & 69.75$\pm$0.54 & 60.48$\pm$2.32 & 61.23$\pm$2.23 & 60.38$\pm$0.38 & 70.70$\pm$2.17 \\
\midrule
\multirow{5}{*}{\textbf{ColBert}}
& Bio.  & 66.03$\pm$0.86 & 62.89$\pm$0.77 & 36.37$\pm$0.21 & 36.13$\pm$0.80 & 70.87$\pm$1.57 \\
& Geo.  & 67.62$\pm$2.22 & 61.84$\pm$0.84 & 35.46$\pm$0.85 & 35.12$\pm$1.66 & 67.06$\pm$0.94 \\
& Chem.  & 65.57$\pm$2.07 & 60.48$\pm$1.54 & 30.79$\pm$0.12 & 31.21$\pm$1.86 & 68.71$\pm$2.43 \\
& Phy.  & 65.88$\pm$2.01 & 60.87$\pm$0.85 & 33.35$\pm$1.44 & 33.96$\pm$1.01 & 69.47$\pm$0.93 \\
& Math. & 66.41$\pm$1.11 & 60.56$\pm$2.32 & 31.64$\pm$1.98 & 33.29$\pm$3.75 & 70.36$\pm$2.11 \\
\midrule
\multirow{5}{*}{\textbf{MAXBAI}}
& Bio.  & 65.56$\pm$1.49 & 63.06$\pm$0.94 & 64.20$\pm$0.97 & 64.91$\pm$1.37 & 71.37$\pm$1.37 \\
& Geo.  & 62.86$\pm$1.37 & 62.08$\pm$1.00 & 63.73$\pm$1.21 & 63.95$\pm$2.05 & 66.42$\pm$1.24 \\
& Chem.  & 62.30$\pm$3.17 & 60.38$\pm$1.56 & 62.87$\pm$0.88 & 64.25$\pm$2.06 & 67.75$\pm$1.65 \\
& Phy.  & 62.64$\pm$3.07 & 61.08$\pm$0.75 & 63.24$\pm$1.45 & 64.10$\pm$0.39 & 67.70$\pm$1.42 \\
& Math. & 62.47$\pm$0.54 & 60.82$\pm$1.80 & 62.77$\pm$2.55 & 62.42$\pm$1.89 & 69.11$\pm$2.06 \\
\midrule
\multirow{5}{*}{\textbf{In-Rank}}
& Bio.  & 71.68$\pm$1.10 & 63.04$\pm$0.86 & 58.56$\pm$0.73 & 57.69$\pm$1.21 & 72.19$\pm$0.90 \\
& Geo.  & 71.39$\pm$0.85 & 62.05$\pm$0.93 & 59.22$\pm$0.52 & 58.09$\pm$0.93 & 67.95$\pm$1.31 \\
& Chem.  & 68.97$\pm$1.82 & 60.43$\pm$1.60 & 57.34$\pm$0.15 & 56.44$\pm$2.64 & 69.44$\pm$2.00 \\
& Phy.  & 69.27$\pm$2.37 & 61.16$\pm$1.04 & 56.85$\pm$1.01 & 57.20$\pm$0.25 & 69.62$\pm$1.04 \\
& Math. & 70.38$\pm$1.12 & 60.53$\pm$2.30 & 56.35$\pm$0.58 & 55.65$\pm$1.49 & 70.82$\pm$0.98 \\
\bottomrule
\end{tabular}
\end{table*}

\begin{table*}[htbp]
\centering
\small
\caption{Performance of different rerankers across various evaluation tasks (AP)}
\label{tab:reranker_tasks_qwen72b_AP}
\begin{tabular}{ccccccc}
\toprule
\textbf{Reranker} & \textbf{Subject} & \textbf{Multi-Hop} & \textbf{NC} & \textbf{CC} & \textbf{SSLI} & \textbf{Base} \\
\midrule
\multirow{5}{*}{\textbf{BCE}}
& Bio.  & 61.40$\pm$1.08 & 59.29$\pm$0.39 & 45.74$\pm$1.20 & 44.97$\pm$1.71 & 69.50$\pm$2.02 \\
& Geo.  & 60.27$\pm$0.98 & 59.95$\pm$0.85 & 44.67$\pm$0.61 & 45.82$\pm$1.57 & 66.07$\pm$0.50 \\
& Chem.  & 58.74$\pm$1.62 & 58.10$\pm$1.69 & 43.67$\pm$1.15 & 43.67$\pm$2.20 & 67.43$\pm$1.46 \\
& Phy.  & 59.06$\pm$4.33 & 58.53$\pm$0.72 & 43.55$\pm$0.66 & 44.79$\pm$1.08 & 68.55$\pm$1.90 \\
& Math. & 59.58$\pm$0.32 & 58.62$\pm$2.08 & 41.63$\pm$0.62 & 41.48$\pm$1.28 & 68.50$\pm$1.87 \\
\midrule
\multirow{5}{*}{\textbf{BGE}}
& Bio.  & 62.01$\pm$0.67 & 60.47$\pm$0.31 & 54.96$\pm$0.96 & 55.52$\pm$1.09 & 73.85$\pm$2.06 \\
& Geo.  & 62.27$\pm$0.59 & 60.55$\pm$0.70 & 53.85$\pm$0.25 & 53.35$\pm$1.89 & 68.75$\pm$0.91 \\
& Chem.  & 59.59$\pm$1.84 & 58.79$\pm$1.73 & 53.49$\pm$0.67 & 53.53$\pm$2.45 & 70.20$\pm$1.74 \\
& Phy.  & 60.38$\pm$3.63 & 59.35$\pm$0.67 & 52.81$\pm$0.91 & 53.44$\pm$0.23 & 70.41$\pm$1.74 \\
& Math. & 61.10$\pm$1.16 & 59.47$\pm$2.04 & 51.67$\pm$1.24 & 51.71$\pm$0.99 & 71.35$\pm$1.88 \\
\midrule
\multirow{5}{*}{\textbf{Jina}}
& Bio.  & 56.86$\pm$0.59 & 58.83$\pm$0.17 & 49.14$\pm$0.76 & 49.01$\pm$0.81 & 70.16$\pm$1.75 \\
& Geo.  & 56.53$\pm$0.24 & 59.23$\pm$0.84 & 51.42$\pm$0.39 & 50.88$\pm$2.03 & 66.22$\pm$1.25 \\
& Chem.  & 56.29$\pm$2.53 & 57.79$\pm$1.95 & 50.37$\pm$0.55 & 49.98$\pm$1.85 & 68.27$\pm$1.93 \\
& Phy.  & 57.21$\pm$3.63 & 58.68$\pm$0.64 & 52.04$\pm$0.78 & 52.48$\pm$0.34 & 68.39$\pm$1.02 \\
& Math. & 56.94$\pm$1.73 & 58.45$\pm$2.16 & 50.30$\pm$1.67 & 49.52$\pm$0.71 & 68.89$\pm$1.07 \\
\midrule
\multirow{5}{*}{\textbf{ListT5}}
& Bio.  & 2.59$\pm$0.11 & 6.53$\pm$0.75 & 1.77$\pm$0.11 & 1.97$\pm$0.51 & 1.97$\pm$0.28 \\
& Geo.  & 2.31$\pm$0.13 & 4.13$\pm$0.66 & 1.30$\pm$0.23 & 1.68$\pm$0.47 & 1.38$\pm$0.14 \\
& Chem.  & 2.38$\pm$0.41 & 5.13$\pm$0.22 & 1.57$\pm$0.20 & 1.45$\pm$0.14 & 1.35$\pm$0.38 \\
& Phy.  & 2.29$\pm$0.23 & 4.91$\pm$0.55 & 1.55$\pm$0.10 & 1.02$\pm$0.10 & 1.40$\pm$0.17 \\
& Math. & 2.63$\pm$0.32 & 4.54$\pm$0.58 & 1.70$\pm$0.06 & 1.58$\pm$0.08 & 1.60$\pm$0.23 \\
\midrule
\multirow{5}{*}{\textbf{MiniLM}}
& Bio.  & 51.83$\pm$0.29 & 59.60$\pm$0.29 & 46.10$\pm$0.43 & 44.80$\pm$0.92 & 70.53$\pm$2.10 \\
& Geo.  & 53.26$\pm$0.53 & 59.71$\pm$0.72 & 44.79$\pm$0.27 & 44.70$\pm$1.18 & 65.88$\pm$1.12 \\
& Chem.  & 50.75$\pm$3.24 & 57.30$\pm$1.53 & 43.98$\pm$1.30 & 43.65$\pm$3.02 & 67.24$\pm$1.79 \\
& Phy.  & 50.48$\pm$3.73 & 58.68$\pm$0.86 & 44.20$\pm$0.89 & 45.17$\pm$0.66 & 67.92$\pm$0.76 \\
& Math. & 50.94$\pm$0.61 & 58.72$\pm$1.99 & 42.28$\pm$0.98 & 42.30$\pm$0.97 & 69.08$\pm$1.58 \\
\midrule
\multirow{5}{*}{\textbf{RankT5}}
& Bio.  & 61.90$\pm$1.41 & 60.23$\pm$0.31 & 52.53$\pm$2.19 & 54.17$\pm$0.57 & 73.06$\pm$1.79 \\
& Geo.  & 61.60$\pm$0.95 & 60.74$\pm$1.28 & 50.09$\pm$4.74 & 48.80$\pm$2.01 & 68.58$\pm$0.89 \\
& Chem.  & 59.90$\pm$2.28 & 59.08$\pm$1.49 & 55.65$\pm$2.77 & 51.90$\pm$2.19 & 70.35$\pm$1.88 \\
& Phy.  & 60.37$\pm$4.01 & 58.99$\pm$0.66 & 57.14$\pm$1.68 & 55.31$\pm$4.42 & 70.11$\pm$0.87 \\
& Math. & 61.14$\pm$0.79 & 59.63$\pm$2.66 & 49.95$\pm$1.53 & 52.55$\pm$3.95 & 71.25$\pm$1.94 \\
\midrule
\multirow{5}{*}{\textbf{SPLADE}}
& Bio.  & 52.07$\pm$0.34 & 58.74$\pm$0.42 & 42.24$\pm$3.32 & 41.91$\pm$2.18 & 66.92$\pm$2.05 \\
& Geo.  & 51.03$\pm$0.29 & 59.39$\pm$1.09 & 41.29$\pm$5.31 & 29.74$\pm$2.25 & 65.15$\pm$0.86 \\
& Chem.  & 51.03$\pm$2.04 & 57.61$\pm$2.24 & 40.89$\pm$2.09 & 36.68$\pm$3.02 & 65.76$\pm$2.48 \\
& Phy.  & 51.47$\pm$3.08 & 58.47$\pm$0.45 & 43.18$\pm$3.87 & 39.23$\pm$6.00 & 65.21$\pm$0.83 \\
& Math. & 52.43$\pm$0.55 & 58.08$\pm$2.47 & 39.96$\pm$3.20 & 37.25$\pm$2.58 & 65.56$\pm$1.86 \\
\midrule
\multirow{5}{*}{\textbf{Twolar}}
& Bio.  & 58.24$\pm$1.07 & 60.03$\pm$0.56 & 59.62$\pm$1.17 & 59.66$\pm$1.66 & 72.08$\pm$1.93 \\
& Geo.  & 57.40$\pm$0.37 & 60.10$\pm$0.76 & 57.06$\pm$0.60 & 55.93$\pm$2.05 & 67.47$\pm$1.10 \\
& Chem.  & 55.93$\pm$2.64 & 58.47$\pm$1.71 & 57.87$\pm$0.66 & 58.77$\pm$2.23 & 69.56$\pm$1.64 \\
& Phy.  & 57.03$\pm$2.73 & 59.25$\pm$0.98 & 59.15$\pm$1.01 & 58.87$\pm$0.26 & 69.79$\pm$0.89 \\
& Math. & 57.49$\pm$1.40 & 59.28$\pm$2.08 & 57.22$\pm$1.67 & 56.35$\pm$1.56 & 70.18$\pm$1.68 \\
\midrule
\multirow{5}{*}{\textbf{ColBert}}
& Bio.  & 54.45$\pm$0.93 & 59.36$\pm$0.37 & 29.70$\pm$0.71 & 29.72$\pm$0.68 & 68.81$\pm$1.74 \\
& Geo.  & 54.94$\pm$0.99 & 59.61$\pm$0.60 & 28.80$\pm$0.64 & 28.54$\pm$1.43 & 65.46$\pm$0.50 \\
& Chem.  & 54.33$\pm$2.67 & 57.87$\pm$1.36 & 26.57$\pm$0.21 & 26.27$\pm$1.03 & 66.98$\pm$2.00 \\
& Phy.  & 54.04$\pm$3.22 & 58.72$\pm$0.83 & 29.35$\pm$0.54 & 29.41$\pm$0.29 & 67.75$\pm$0.74 \\
& Math. & 54.72$\pm$0.52 & 58.77$\pm$2.29 & 28.68$\pm$1.12 & 27.38$\pm$2.72 & 68.47$\pm$1.42 \\
\midrule
\multirow{5}{*}{\textbf{MAXBAI}}
& Bio.  & 56.42$\pm$1.54 & 59.10$\pm$0.87 & 60.47$\pm$0.82 & 62.19$\pm$1.31 & 68.65$\pm$1.74 \\
& Geo.  & 55.48$\pm$1.64 & 59.53$\pm$0.90 & 59.37$\pm$1.02 & 59.80$\pm$1.88 & 64.12$\pm$0.69 \\
& Chem.  & 55.19$\pm$2.54 & 57.88$\pm$2.03 & 60.45$\pm$0.97 & 61.35$\pm$1.98 & 66.37$\pm$2.12 \\
& Phy.  & 55.36$\pm$3.24 & 58.66$\pm$0.84 & 60.45$\pm$0.38 & 60.62$\pm$0.50 & 66.64$\pm$1.37 \\
& Math. & 54.07$\pm$0.50 & 58.61$\pm$1.79 & 60.22$\pm$3.21 & 59.45$\pm$2.20 & 67.56$\pm$2.02 \\
\midrule
\multirow{5}{*}{\textbf{In-Rank}}
& Bio.  & 61.09$\pm$0.33 & 59.60$\pm$0.42 & 52.46$\pm$0.11 & 53.13$\pm$0.87 & 71.43$\pm$2.11 \\
& Geo.  & 60.45$\pm$0.67 & 60.37$\pm$0.75 & 54.23$\pm$0.30 & 52.96$\pm$2.13 & 66.94$\pm$0.82 \\
& Chem.  & 58.76$\pm$2.08 & 58.32$\pm$1.77 & 52.45$\pm$0.54 & 52.36$\pm$2.45 & 68.81$\pm$1.73 \\
& Phy.  & 60.44$\pm$4.09 & 59.04$\pm$0.83 & 53.60$\pm$0.68 & 53.85$\pm$0.76 & 68.43$\pm$0.76 \\
& Math. & 60.65$\pm$0.79 & 59.24$\pm$2.03 & 52.61$\pm$1.04 & 50.10$\pm$0.62 & 69.75$\pm$1.36 \\
\bottomrule
\end{tabular}
\end{table*}

\begin{table*}[htbp]
\centering
\small
\caption{Performance of different rerankers across various evaluation tasks (MRR)}
\label{tab:reranker_tasks_qwen72b_MRR}
\begin{tabular}{ccccccc}
\toprule
\textbf{Reranker} & \textbf{Subject} & \textbf{Multi-Hop} & \textbf{NC} & \textbf{CC} & \textbf{SSLI} & \textbf{Base} \\
\midrule
\multirow{5}{*}{\textbf{BCE}}
& Bio.  & 0.63$\pm$0.01 & 0.60$\pm$0.01 & 0.46$\pm$0.01 & 0.45$\pm$0.02 & 0.71$\pm$0.02 \\
& Geo.  & 0.63$\pm$0.01 & 0.61$\pm$0.01 & 0.45$\pm$0.01 & 0.46$\pm$0.02 & 0.68$\pm$0.01 \\
& Chem.  & 0.59$\pm$0.02 & 0.59$\pm$0.02 & 0.43$\pm$0.01 & 0.43$\pm$0.02 & 0.69$\pm$0.01 \\
& Phy.  & 0.60$\pm$0.04 & 0.59$\pm$0.01 & 0.43$\pm$0.00 & 0.44$\pm$0.01 & 0.70$\pm$0.01 \\
& Math. & 0.62$\pm$0.01 & 0.59$\pm$0.02 & 0.42$\pm$0.01 & 0.41$\pm$0.01 & 0.70$\pm$0.01 \\
\midrule
\multirow{5}{*}{\textbf{BGE}}
& Bio.  & 0.64$\pm$0.00 & 0.61$\pm$0.00 & 0.56$\pm$0.01 & 0.56$\pm$0.01 & 0.76$\pm$0.02 \\
& Geo.  & 0.65$\pm$0.00 & 0.61$\pm$0.01 & 0.54$\pm$0.00 & 0.54$\pm$0.02 & 0.71$\pm$0.01 \\
& Chem.  & 0.60$\pm$0.01 & 0.59$\pm$0.02 & 0.53$\pm$0.01 & 0.53$\pm$0.02 & 0.72$\pm$0.02 \\
& Phy.  & 0.61$\pm$0.04 & 0.60$\pm$0.01 & 0.53$\pm$0.01 & 0.54$\pm$0.00 & 0.72$\pm$0.02 \\
& Math. & 0.62$\pm$0.01 & 0.60$\pm$0.02 & 0.52$\pm$0.01 & 0.52$\pm$0.01 & 0.73$\pm$0.02 \\
\midrule
\multirow{5}{*}{\textbf{Jina}}
& Bio.  & 0.58$\pm$0.01 & 0.60$\pm$0.00 & 0.50$\pm$0.01 & 0.50$\pm$0.01 & 0.72$\pm$0.02 \\
& Geo.  & 0.59$\pm$0.00 & 0.60$\pm$0.01 & 0.52$\pm$0.00 & 0.52$\pm$0.02 & 0.68$\pm$0.02 \\
& Chem.  & 0.57$\pm$0.03 & 0.58$\pm$0.02 & 0.51$\pm$0.01 & 0.50$\pm$0.02 & 0.70$\pm$0.02 \\
& Phy.  & 0.59$\pm$0.04 & 0.59$\pm$0.01 & 0.52$\pm$0.01 & 0.53$\pm$0.00 & 0.70$\pm$0.01 \\
& Math. & 0.58$\pm$0.02 & 0.59$\pm$0.02 & 0.51$\pm$0.02 & 0.50$\pm$0.00 & 0.70$\pm$0.01 \\
\midrule
\multirow{5}{*}{\textbf{ListT5}}
& Bio.  & 0.03$\pm$0.00 & 0.07$\pm$0.01 & 0.02$\pm$0.00 & 0.02$\pm$0.01 & 0.02$\pm$0.00 \\
& Geo.  & 0.02$\pm$0.00 & 0.04$\pm$0.01 & 0.01$\pm$0.00 & 0.02$\pm$0.01 & 0.01$\pm$0.00 \\
& Chem.  & 0.02$\pm$0.00 & 0.05$\pm$0.00 & 0.02$\pm$0.00 & 0.01$\pm$0.00 & 0.01$\pm$0.00 \\
& Phy.  & 0.02$\pm$0.00 & 0.05$\pm$0.01 & 0.02$\pm$0.00 & 0.01$\pm$0.00 & 0.01$\pm$0.00 \\
& Math. & 0.03$\pm$0.00 & 0.05$\pm$0.01 & 0.02$\pm$0.00 & 0.02$\pm$0.00 & 0.02$\pm$0.00 \\
\midrule
\multirow{5}{*}{\textbf{MiniLM}}
& Bio.  & 0.53$\pm$0.01 & 0.60$\pm$0.00 & 0.47$\pm$0.00 & 0.45$\pm$0.01 & 0.73$\pm$0.02 \\
& Geo.  & 0.55$\pm$0.01 & 0.61$\pm$0.01 & 0.45$\pm$0.00 & 0.45$\pm$0.01 & 0.68$\pm$0.01 \\
& Chem.  & 0.52$\pm$0.04 & 0.58$\pm$0.01 & 0.44$\pm$0.01 & 0.43$\pm$0.03 & 0.69$\pm$0.02 \\
& Phy.  & 0.52$\pm$0.04 & 0.59$\pm$0.01 & 0.44$\pm$0.01 & 0.45$\pm$0.00 & 0.69$\pm$0.01 \\
& Math. & 0.53$\pm$0.01 & 0.59$\pm$0.02 & 0.43$\pm$0.01 & 0.42$\pm$0.01 & 0.71$\pm$0.01 \\
\midrule
\multirow{5}{*}{\textbf{RankT5}}
& Bio.  & 0.64$\pm$0.01 & 0.61$\pm$0.00 & 0.53$\pm$0.02 & 0.55$\pm$0.00 & 0.75$\pm$0.02 \\
& Geo.  & 0.64$\pm$0.01 & 0.61$\pm$0.01 & 0.52$\pm$0.05 & 0.49$\pm$0.02 & 0.71$\pm$0.01 \\
& Chem.  & 0.61$\pm$0.02 & 0.60$\pm$0.01 & 0.55$\pm$0.03 & 0.52$\pm$0.02 & 0.72$\pm$0.02 \\
& Phy.  & 0.62$\pm$0.04 & 0.59$\pm$0.01 & 0.57$\pm$0.02 & 0.55$\pm$0.04 & 0.72$\pm$0.01 \\
& Math. & 0.63$\pm$0.01 & 0.60$\pm$0.03 & 0.51$\pm$0.02 & 0.53$\pm$0.03 & 0.73$\pm$0.02 \\
\midrule
\multirow{5}{*}{\textbf{SPLADE}}
& Bio.  & 0.53$\pm$0.01 & 0.59$\pm$0.01 & 0.42$\pm$0.03 & 0.42$\pm$0.02 & 0.69$\pm$0.02 \\
& Geo.  & 0.53$\pm$0.00 & 0.60$\pm$0.01 & 0.42$\pm$0.06 & 0.30$\pm$0.03 & 0.67$\pm$0.01 \\
& Chem.  & 0.52$\pm$0.02 & 0.58$\pm$0.02 & 0.40$\pm$0.02 & 0.35$\pm$0.03 & 0.67$\pm$0.03 \\
& Phy.  & 0.52$\pm$0.04 & 0.59$\pm$0.00 & 0.42$\pm$0.04 & 0.38$\pm$0.06 & 0.66$\pm$0.01 \\
& Math. & 0.53$\pm$0.01 & 0.59$\pm$0.02 & 0.40$\pm$0.04 & 0.37$\pm$0.02 & 0.67$\pm$0.01 \\
\midrule
\multirow{5}{*}{\textbf{Twolar}}
& Bio.  & 0.60$\pm$0.01 & 0.61$\pm$0.01 & 0.61$\pm$0.01 & 0.61$\pm$0.02 & 0.74$\pm$0.02 \\
& Geo.  & 0.59$\pm$0.00 & 0.61$\pm$0.01 & 0.58$\pm$0.01 & 0.57$\pm$0.02 & 0.69$\pm$0.01 \\
& Chem.  & 0.56$\pm$0.03 & 0.59$\pm$0.02 & 0.58$\pm$0.01 & 0.59$\pm$0.02 & 0.71$\pm$0.02 \\
& Phy.  & 0.58$\pm$0.03 & 0.60$\pm$0.01 & 0.60$\pm$0.01 & 0.59$\pm$0.00 & 0.71$\pm$0.01 \\
& Math. & 0.59$\pm$0.02 & 0.60$\pm$0.02 & 0.58$\pm$0.02 & 0.57$\pm$0.02 & 0.72$\pm$0.02 \\
\midrule
\multirow{5}{*}{\textbf{ColBert}}
& Bio.  & 0.55$\pm$0.01 & 0.60$\pm$0.00 & 0.29$\pm$0.01 & 0.29$\pm$0.01 & 0.71$\pm$0.02 \\
& Geo.  & 0.57$\pm$0.01 & 0.60$\pm$0.01 & 0.29$\pm$0.01 & 0.29$\pm$0.02 & 0.67$\pm$0.01 \\
& Chem.  & 0.55$\pm$0.03 & 0.58$\pm$0.01 & 0.25$\pm$0.00 & 0.25$\pm$0.01 & 0.68$\pm$0.02 \\
& Phy.  & 0.55$\pm$0.04 & 0.59$\pm$0.01 & 0.29$\pm$0.00 & 0.29$\pm$0.00 & 0.69$\pm$0.01 \\
& Math. & 0.56$\pm$0.01 & 0.60$\pm$0.02 & 0.29$\pm$0.01 & 0.27$\pm$0.03 & 0.70$\pm$0.01 \\
\midrule
\multirow{5}{*}{\textbf{MAXBAI}}
& Bio.  & 0.58$\pm$0.02 & 0.60$\pm$0.01 & 0.62$\pm$0.00 & 0.64$\pm$0.01 & 0.71$\pm$0.02 \\
& Geo.  & 0.58$\pm$0.02 & 0.60$\pm$0.01 & 0.61$\pm$0.01 & 0.61$\pm$0.02 & 0.66$\pm$0.01 \\
& Chem.  & 0.56$\pm$0.02 & 0.58$\pm$0.02 & 0.61$\pm$0.01 & 0.62$\pm$0.02 & 0.68$\pm$0.02 \\
& Phy.  & 0.57$\pm$0.04 & 0.59$\pm$0.01 & 0.61$\pm$0.00 & 0.61$\pm$0.01 & 0.68$\pm$0.01 \\
& Math. & 0.56$\pm$0.01 & 0.59$\pm$0.02 & 0.61$\pm$0.03 & 0.61$\pm$0.02 & 0.69$\pm$0.02 \\
\midrule
\multirow{5}{*}{\textbf{In-Rank}}
& Bio.  & 0.63$\pm$0.00 & 0.60$\pm$0.00 & 0.53$\pm$0.00 & 0.54$\pm$0.01 & 0.74$\pm$0.02 \\
& Geo.  & 0.62$\pm$0.01 & 0.61$\pm$0.01 & 0.55$\pm$0.00 & 0.54$\pm$0.02 & 0.69$\pm$0.01 \\
& Chem.  & 0.60$\pm$0.02 & 0.59$\pm$0.02 & 0.52$\pm$0.00 & 0.52$\pm$0.02 & 0.70$\pm$0.02 \\
& Phy.  & 0.62$\pm$0.04 & 0.59$\pm$0.01 & 0.54$\pm$0.01 & 0.54$\pm$0.01 & 0.70$\pm$0.01 \\
& Math. & 0.62$\pm$0.01 & 0.60$\pm$0.02 & 0.53$\pm$0.01 & 0.50$\pm$0.00 & 0.71$\pm$0.01 \\
\bottomrule
\end{tabular}
\end{table*}






\definecolor{lightgray}{gray}{0.95}
\begin{figure*}[htbp]
\centering
\small
\begin{tcolorbox}[colback=gray!5!white, colframe=gray!80!black, title=A Counterfactual Synthetic (Question-Context-Answer) Sample.]
\underline{\textbf{Question:}} What are two viable large-scale energy storage technologies?

\underline{\textbf{Golden Answer:}} Underground Hydrogen Storage (UHS) and Compressed Air Energy storage (CAES).

\underline{\textbf{Context:}}
\begin{itemize}
  \item \textit{Passage 1}: Underground hydrogen storage (UHS) and compressed air energy storage (CAES) are two viable large-scale energy storage technologies for mitigating the intermittency of wind and solar power. Therefore, it is meaningful to compare the properties of hydrogen and air with typical thermodynamic storage processes. This study employs a multi-physical coupling model to compare the operations of CAES and UHS, integrating gas thermodynamics within caverns, thermal conduction, and mechanical deformation around rock caverns. Gas thermodynamic responses are validated using additional simulations and the field test data. Temperature and pressure variations of air and hydrogen within rock caverns exhibit similarities under both adiabatic and diabatic simulation modes. Hydrogen reaches higher temperature and pressure following gas charging stage compared to air, and the ideal gas assumption may lead to overestimation of gas temperature and pressure. Unlike steel lining of CAES, the sealing layer (fibre-reinforced plastic FRP) in UHS is prone to deformation but can effectively mitigates stress in the sealing layer. In CAES, the first principal stress on the surface of the sealing layer and concrete lining is tensile stress, whereas UHS exhibits compressive stress in the same areas. Our present research can provide references for the selection of energy storage methods.
  \item \textit{Passage 2}: Intensive increases in electrical energy storage are being driven by electric vehicles (EVs), smart grids, intermittent renewable energy, and decarbonization of the energy economy. Advanced lithium–sulfur batteries (LSBs) are among the most promising candidates, especially for EVs and grid-scale energy storage applications. In this topical review, the recent progress and perspectives of practical LSBs are reviewed and discussed; the challenges and solutions for these LSBs are analyzed and proposed for future practical and large-scale energy storage applications. Major challenges for the shuttle effect, reaction kinetics, and anodes are specifically addressed, and solutions are provided on the basis of recent progress in electrodes, electrolytes, binders, interlayers, conductivity, electrocatalysis, artificial SEI layers, etc. The characterization strategies (including in situ ones) and practical parameters (e.g., cost-effectiveness, battery management/modeling, environmental adaptability) are assessed for crucial automotive/stationary large-scale energy storage applications (i.e., EVs and grid energy storage). This topical review will give insights into the future development of promising Li–S batteries toward practical applications, including EVs and grid storage.
  \item \textit{Passage 3}: To support increasing renewable capacity for a net-zero future, energy storage will play a key role in maintaining grid stability. In this paper, all current and near-future energy storage technologies are compared for three different scenarios: (1) fixed electricity buy-in price, (2) market-based electricity buy-in price, and (3) energy storage integrated into a fully renewable electricity system. In the first part of this study, an algorithm is devised to simulate strategic buy-in of electricity for energy storage. This analysis yields a qualitative decision-making tool for a given energy storage duration and size. Building upon the first part’s findings, an integration study gives insight into expected power prices and expected storage size in a typical northwestern European fully renewable energy system. The integration study shows significant need for electricity storage with durations spanning from one to several days, typically around 40 h. Pumped Hydro Storage and Pumped Thermal storage surface as the best options. The overall levelized costs of storage are expected to be in the USD 200–500/MWh range. Integration of storage with renewables can yield a system-levelized cost of electricity of about USD 150/MWh. Allowing flexibility in demand may lower the overall system-levelized cost of electricity to USD 100/MWh.","Climate change mitigation requires the large-scale deployment of carbon capture and storage (CCS). Recent plans indicate an eight-fold increase in CCS capacity by 2030, yet the feasibility of CCS expansion is debated. Using historical growth of CCS and other policy-driven technologies, we show that if plans double between 2023 and 2025 and their failure rates decrease by half, CCS could reach 0.37 GtCO
  \item \dots
  \item \textit{Passage 99}: Two large-scale energy storage technologies with limited safety: Sodium-ion batteries, which are known for their safety, and lithium-ion batteries, which have a higher risk of thermal runaway, are both unsuitable for use in a safe energy storage system.
  \item \textit{Passage 100}: Two large-scale energy storage technologies with limited environmental sustainability: Lead-acid batteries, which are known for their environmental sustainability, and lithium-ion batteries, which have a higher environmental impact due to their production and disposal, are both unsuitable for use in an environmentally sustainable energy storage system. 
  
\end{itemize}
\end{tcolorbox}
\caption{A Counterfactual synthetic (Question, Context, Answer) Sample}
\label{fig:dataste-sample-1}
\end{figure*}

\definecolor{lightgray}{gray}{0.95}
\begin{figure*}[htbp]
\centering
\small
\begin{tcolorbox}[colback=gray!5!white, colframe=gray!80!black, title=A SSLI Synthetic (Question-Context-Answer) Sample.]
\underline{\textbf{Question:}} What is the most lethal melanoma subtype?

\underline{\textbf{Golden Answer:}} uveal melanomas

\underline{\textbf{Context:}}
\begin{itemize}
  \item \textit{Passage 1}: Activating mutations in GNAQ/GNA11 occur in over 90\% of uveal melanomas (UMs), the most lethal melanoma subtype; however, targeting these oncogenes has proven challenging and inhibiting their downstream effectors show limited clinical efficacy. Here, we performed genome-scale CRISPR screens along with computational analyses of cancer dependency and gene expression datasets to identify the inositol-metabolizing phosphatase INPP5A as a selective dependency in GNAQ/11-mutant UM cells in vitro and in vivo. Mutant cells intrinsically produce high levels of the second messenger inositol 1,4,5 trisphosphate (IP3) that accumulate upon suppression of INPP5A, resulting in hyperactivation of IP3-receptor signaling, increased cytosolic calcium and p53-dependent apoptosis. Finally, we show that GNAQ/11-mutant UM cells and patients’ tumors exhibit elevated levels of IP4, a biomarker of enhanced IP3 production; these high levels are abolished by GNAQ/11 inhibition and correlate with sensitivity to INPP5A depletion. Our findings uncover INPP5A as a synthetic lethal vulnerability and a potential therapeutic target for GNAQ/11-mutant-driven cancers.
  \item \textit{Passage 2}: Background Although anti-Program-Death-1 (PD-1) is a standard adjuvant therapy for patients with resected melanoma. We hypothesize that there are discrepancies of survival, recurrence pattern and toxicity to adjuvant PD-1 between different ethnicities and melanoma subtypes. Objective/methodsWe performed a multicenter cohort study incorporating 6 independent institutions in Australia, China, Japan, and US.The primary outcomes were RFS and OS.Secondary outcomes were disease recurrence patterns and toxicities. ResultsIn total 534 patients were included.East-Asian/Hispanic/African had significantly poorer RFS/OS.Non-acral-cutaneous/unknown primary (NAC/UP) had the best RFS/OS, followed by acral; mucosal the poorest.Within the NAC/UP subtypes, East-Asian/Hispanic/African had significantly poorer RFS/OS than Caucasian.In the multivariate analysis incorporating ethnicity/melanoma-subtype/age/sex/stage/LDH/BRAFmutation/adjuvant-radiotherapy, East-Asian/Hispanic/African had independently significantly poorer outcomes (RFS: HR1.71 95\%CI 1.19-2.44;OS: HR2.34, 95\%CI 1.39-3.95),as was mucosal subtype (RFS: HR3.25, 95\%CI 2.04-5.17;OS: HR3.20, 95\%CI 1.68-6.08).Mucosal melanoma was an independent risk factor for distant-metastasis, especially liver metastasis.East-Asian/Hispanic/African had significantly lower incidence of GI/musculoskeletal/respiratory/other-rare-type-toxicities; but higher incidences of liver toxicities. Limitations retrospective study.Conclusions Ethnicity and melanoma subtype are associated with survival and recurrence pattern in melanoma patients treated with adjuvant anti-PD-1.Toxicity profile differs by ethnicity, and may require a precision toxicity surveillance strategy.
  \item \textit{Passage 3}: Melanoma is the third most common type of skin cancer, characterized by its heterogeneity and propensity to metastasize to distant organs. Melanoma is a heterogeneous tumor, composed of genetically divergent subpopulations, including a small fraction of melanoma-initiating cancer stem-like cells (CSCs) and many non-cancer stem cells (non-CSCs). CSCs are characterized by their unique surface proteins associated with aberrant signaling pathways with a causal or consequential relationship with tumor progression, drug resistance, and recurrence. Melanomas also harbor significant alterations in functional genes (BRAF, CDKN2A, NRAS, TP53, and NF1). Of these, the most common are the BRAF and NRAS oncogenes, with 50\% of melanomas demonstrating the BRAF mutation (BRAF
  \item \textit{Passage 4}: Melanoma, the deadliest form of skin cancer, poses a significant clinical challenge for the development of effective treatments. Conventional in vivo animal studies have shown limited translational relevance to humans, raising strength to pre-clinical models for melanoma research. This review provides an in-depth analysis of alternative pre-clinical models including in vitro and ex vivo platforms such as reconstructed skin, spheroids, organoids, organotypic models, skin-on-a-chip, and bioprinting. Through a comprehensive analysis, the specific attributes, advantages, and limitations of each model are elucidated. It discusses the points related to the uniqueness advantages, from capturing complex interactions between melanoma cells and their microenvironment to enabling high-throughput drug screening and personalized medicine approaches. This review is structured covering firstly the roadmap to identify the co-occurrence of discovering new melanoma treatments and the development of its models, secondly it covers a comparative between the most used models followed by a section discussing each of them: the in vitro and ex vivo models. It intends to serve as an asset for researchers of melanoma field and clinicians involved in melanoma therapy, offering insights into the diverse preclinical models available for optimizing their integration into the translational pipeline. 
  \item \dots
  \item \textit{Passage 99}: Deceptive: The most lethal melanoma subtype is often misdiagnosed as a harmless freckle, making early detection challenging.
  \item \textit{Passage 100}: Perplexing: It's a common misconception that the most lethal melanoma subtype is the one that appears as a mole.
  
\end{itemize}
\end{tcolorbox}
\caption{A SSLI synthetic (Question, Context, Answer) Sample}
\label{fig:dataste-sample-2}
\end{figure*}

\definecolor{lightgray}{gray}{0.95}
\begin{figure*}[htbp]
\centering
\small
\begin{tcolorbox}[colback=gray!5!white, colframe=gray!80!black, title=A Multi-Hop Synthetic (Question-Context-Answer) Sample.]
\underline{\textbf{Question:}} What are the isoforms categories of genes that control activity of gene transcription?

\underline{\textbf{Golden Answer:}} rewirers and negative regulators

\underline{\textbf{Context:}}
\begin{itemize}
  \item \textit{Passage 1}: Most human Transcription factors (TFs) genes encode multiple protein isoforms differing in DNA binding domains, effector domains, or other protein regions. The global extent to which this results in functional differences between isoforms remains unknown. Here, we systematically compared 693 isoforms of 246 TF genes, assessing DNA binding, protein binding, transcriptional activation, subcellular localization, and condensate formation. Relative to reference isoforms, two-thirds of alternative TF isoforms exhibit differences in one or more molecular activities, which often could not be predicted from sequence. We observed two primary categories of alternative TF isoforms: \"rewirers\" and \"negative regulators\", both of which were associated with differentiation and cancer. Our results support a model wherein the relative expression levels of, and interactions involving, TF isoforms add an understudied layer of complexity to gene regulatory networks, demonstrating the importance of isoform-aware characterization of TF functions and providing a rich resource for further studies. 
  \item \textit{Passage 2}: More than 1600 human transcription factors orchestrate the transcriptional machinery to control gene expression and cell fate. Their function is conveyed through intrinsically disordered regions (IDRs) containing activation or repression domains but lacking quantitative structural ensemble models prevents their mechanistic decoding. Here we integrate single-molecule FRET and NMR spectroscopy with molecular simulations showing that DNA binding can lead to complex changes in the IDR ensemble and accessibility. The C-terminal IDR of pioneer factor Sox2 is highly disordered but its conformational dynamics are guided by weak and dynamic charge interactions with the folded DNA binding domain. Both DNA and nucleosome binding induce major rearrangements in the IDR ensemble without affecting DNA binding affinity. Remarkably, interdomain interactions are redistributed in complex with DNA leading to variable exposure of two activation domains critical for transcription. Charged intramolecular interactions allowing for dynamic redistributions may be common in transcription factors and necessary for sensitive tuning of structural ensembles.
  \item \textit{Passage 3}: Transcription factors (TFs) control specificity and activity of gene transcription, but whether a relationship between these two features exists is unclear. Here we provide evidence for an evolutionary trade-off between the activity and specificity in human TFs encoded as submaximal dispersion of aromatic residues in their intrinsically disordered protein regions. We identified approximately 500 human TFs that encode short periodic blocks of aromatic residues in their intrinsically disordered regions, resembling imperfect prion-like sequences. Mutation of periodic aromatic residues reduced transcriptional activity, whereas increasing the aromatic dispersion of multiple human TFs enhanced transcriptional activity and reprogramming efficiency, promoted liquid–liquid phase separation in vitro and more promiscuous DNA binding in cells. Together with recent work on enhancer elements, these results suggest an important evolutionary role of suboptimal features in transcriptional control. We propose that rational engineering of amino acid features that alter phase separation may be a strategy to optimize TF-dependent processes, including cellular reprogramming.
  \item \dots
  \item \textit{Passage 99}: Pioneer transcription factors (TFs) exhibit a specialized ability to bind to and open closed chromatin, facilitating engagement by other regulatory factors involved in gene activation or repression. Chemical probes are lacking for pioneer TFs, which has hindered their mechanistic investigation in cells. Here, we report the chemical proteomic discovery of electrophilic small molecules that stereoselectively and site-specifically bind the pioneer TF, FOXA1, at a cysteine (C258) within the forkhead DNA-binding domain. We show that these covalent ligands react with FOXA1 in a DNA-dependent manner and rapidly remodel its pioneer activity in prostate cancer cells reflected in redistribution of FOXA1 binding across the genome and directionally correlated changes in chromatin accessibility. Motif analysis supports a mechanism where the covalent ligands relax the canonical DNA binding preference of FOXA1 by strengthening interactions with suboptimal ancillary sequences in predicted proximity to C258. Our findings reveal a striking plasticity underpinning the pioneering function of FOXA1 that can be controlled by small molecules.
  \item \textit{Passage 100}: Alternative transcription start site usage (ATSS) is a widespread regulatory strategy that enables genes to choose between multiple genomic loci for initiating transcription. This mechanism is tightly controlled during development and is often altered in disease states. In this review, we examine the growing evidence highlighting a role for transcription start sites (TSSs) in the regulation of mRNA isoform selection during and after transcription. We discuss how the choice of transcription initiation sites influences RNA processing and the importance of this crosstalk for cell identity and organism function. We also speculate on possible mechanisms underlying the integration of transcriptional and post-transcriptional processes. 
  
\end{itemize}
\end{tcolorbox}
\caption{A multi-hop synthetic (Question, Context, Answer) sample.}
\label{fig:dataste-sample-3}
\end{figure*}

\definecolor{lightgray}{gray}{0.95}
\begin{figure*}[htbp]
\centering
\small
\begin{tcolorbox}[colback=gray!5!white, colframe=gray!80!black, title=A Random Mix Synthetic (Question-Context-Answer) Sample (NC).]
\underline{\textbf{Question:}} What are perovskite optoelectronic devices utilizing lead-free perovskites?

\underline{\textbf{Golden Answer:}} solar cells, LEDs, and near-infrared photodetectors.

\underline{\textbf{Context:}}
\begin{itemize}
  \item \textit{Passage 1}: In modern technology, optoelectronics plays a pivotal role, finding applications in diverse areas like solar cells, light-emitting diodes (LEDs), and photodetectors. Perovskite materials, known for their exceptional optical properties, are gaining attraction due to their light weight, flexibility, and ease of production. The expanding markets of wearable electronics and flexible displays have underscored the importance of developing flexible perovskite optoelectronic devices. However, the prevalent use of lead in high-performance perovskite devices poses significant environmental and health risks, casting doubt on their commercial future. This review commences with examining lead hazards, followed by a discussion on how first-principles calculations aid in designing lead-free perovskites. We survey the synthesized lead-free perovskites and explore their properties. The focus then shifts to the latest advancements in flexible optoelectronic devices utilizing lead-free perovskites, including solar cells, LEDs, and near-infrared photodetectors. Additionally, we explore the role of TCAD (Technology Computer-Aided Design) in simulating and optimizing these devices, highlighting its impact on device design and efficiency.
  \item \textit{Passage 2}: In the exploration of perovskite materials devoid of lead and appropriate for capturing solar energy, a recent finding has surfaced concerning Cs$_2$ZrCl$_6$.
  \item \textit{Passage 3}: In the pursuit of commercializing electronic and optoelectronic devices, researchers have turned their attention to non-toxic inorganic cubic metal halide perovskites. This study focuses on novel lead-free compounds—specifically AlGeX$_3$ (where X = Cl, Br, I) for potential use in next-generation energy-harvesting devices.
  \item \textit{Passage 4}: In recent years, halide perovskite materials have become widely used in solar cells, photovoltaics, and LEDs, as well as photocatalysis. Lead-free perovskite Cs.
  \item \textit{Passage 5}: Abstract Science teacher identity significantly influences teacher professional development, practices, and attitudes, which in turn impacts student learning outcomes. With an increased number of studies on science teacher identity over the past two decades, there is a need for a scoping literature review that holistically maps the current state of science teacher identity research and identifies future research directions. This scoping literature review identified 48 empirical articles on science teacher identity, published from 2000 to 2023, in peer-reviewed journals and examined the studies’ (a) characteristics; (b) theoretical frameworks on identity; (c) definitions of science teacher identity; and (d) major findings. Specifically, there is a need for precise conceptualizations and definitions of science teacher identity; this clarity will facilitate valid, reliable, and fair instruments to capture the relatively stable facets of science teacher identity at a given moment in a given context in order to longitudinally track science teacher identity development. This scoping review identifies both progress and gaps in the current literature and future directions for synergistic, cross-cultural international research on science teacher identity.
  \item \dots
  \item \textit{Passage 99}: This study conducts a bibliometric analysis using the Web of Science database on 1,950 articles published in the journal Mindfulness from 2012 to 2022. By constructing a knowledge graph, the research delineates the evolution, stages of development, and emerging trends in the field of mindfulness. Significant growth in the annual publication volume has been observed since 2012, with the research progression segmented into three distinct phases. The United States has emerged as a pivotal contributor to the field, dominating in terms of publication volume, researcher involvement, and institutional contributions. Through the application of keyword co-occurrence and reference co-citation analysis, five principal clusters were identified, focusing on mindfulness, meditation, depression, stress, and self-compassion, underscoring these as focal research areas. Furthermore, the exploration of mindfulness within the educational sphere in Taiwan is still nascent, signaling a critical need for bolstered research support in diverse thematic domains.
  \item \textit{Passage 100}: Stroke is a cerebral vascular disease characterized by the death of brain tissue that occurs due to reduced blood and oxygen flow to the brain. Ischemic stroke is associated with diabetes mellitus, therefore it is important to identify the risk factors that cause stroke and DM by diagnostic cause of the disease. This study aimed to classify and compare accuracy tests on medical record data sets for stroke and DM. This study analyzed the diagnosis of stroke and DM using Decision Tree. The risk factors consisted of gender, age, blood pressure, nutritional status, smoking, history of DM, and history of hypertension. The results of the analysis using the Decision Tree method showed that the accuracy rate was 86.67\%, which means that the modeling has a good level of correctness of the prediction results. We conclude that the Decision Tree method was an accurate method for detecting stroke and DM.

\end{itemize}
\end{tcolorbox}
\caption{A random mix synthetic (Question, Context, Answer) Sample (NC)}
\label{fig:dataste-sample-4}
\end{figure*}

\end{document}